\documentclass[acmlarge,screen]{acmart}
\AtBeginDocument{%
  }

\setcopyright{acmlicensed}
\copyrightyear{2026}
\acmYear{2026}
\acmDOI{XXXXXXX.XXXXXXX}

\acmJournal{CSUR}
\acmVolume{58}
\acmNumber{4}
\acmArticle{111}
\acmMonth{8}

\usepackage[ruled,vlined]{algorithm2e}

\begin{document}

\title{A Tutorial on World Models and Physical AI}

\author{Il-Seok Oh}
\authornote{Both authors contributed equally to this research.}
\email{isoh@jbnu.ac.kr}
\orcid{0000-0002-8823-0438}
\affiliation{%
  \institution{Department of Computer Science and Artificial Intelligence/CAIIT}
  \city{Jeonju}
  \state{Jeonbuk}
  \country{South Korea}
}

\renewcommand{\shortauthors}{Il-Seok Oh}

\begin{abstract}
  World modeling is emerging as a central principle for building intelligent systems capable of prediction, reasoning, and decision making. A central distinction can be drawn between explicit world models, which learn structured dynamics for rollout-based reasoning and planning, and implicit world models, which encode predictive structure within scalable learned representations. These complementary paradigms provide a foundation for physical AI in domains such as robotics and autonomous driving, enabling intelligence beyond reactive control under real-world constraints. Recent foundation models further suggest a pathway toward unified systems integrating perception, prediction, and action. Despite rapid progress, major challenges remain in hierarchical reasoning, long-horizon planning, and autonomous goal formation, which are critical for advancing toward artificial general intelligence. This tutorial presents a coherent framework in which diverse world modeling approaches are unified through shared predictive structure and differentiated by how such structure is represented and exploited.
\end{abstract}



\received{20 February 2026}
\received[revised]{12 March 2026}
\received[accepted]{5 June 2026}

\maketitle

\section{Introduction}
Humans live in a world governed by structured regularities, spanning physical, social, and causal dynamics. To function effectively within such an environment, they continuously anticipate the consequences of their actions \cite{Johnson-Laird1983}. This predictive ability is grounded in an internal world model formed through experience. By leveraging this model, humans can imagine future outcomes, avoid potentially dangerous actions, and adapt to their environment with limited trial and error. Thus, intelligence fundamentally relies not on purely reactive behavior, but on prediction and judgment based on an internalized model of the world’s structure and dynamics.

Reinforcement learning (RL) has long served as a central framework for sequential decision making in artificial intelligence, formalizing how an agent learns to act through trial-and-error interaction with an environment \cite{Sutton2018}. In highly complex and uncertain environments, effective decision making requires more than reacting to the current state; agents must anticipate the future consequences of their actions and reason over long horizons. While RL provides a principled formulation for such problems, classical approaches typically rely on direct interaction with the environment and task-specific reward signals. As a result, they often require large amounts of data, exhibit limited generalization across tasks, and struggle in long-horizon or safety-critical settings where exhaustive exploration is impractical. Effective solutions, therefore, require mechanisms that enable agents to reason about long-term consequences, imagine future trajectories, and reduce reliance on costly real-world interaction.

World models have emerged as a key concept for addressing these limitations by equipping agents with an internal representation of the environment's dynamics \cite{Ding2025}. By learning how the world evolves as a consequence of actions, an agent can imagine future rollouts, evaluate alternative behaviors, and plan without relying exclusively on real-world interaction. This idea has deep roots in model-based reinforcement learning \cite{Moerland2023}, where explicit transition models were learned to support planning and simulation. More recent advances have extended this paradigm by learning rich latent dynamics models that scale to high-dimensional observations, enabling agents to perform imagination, counterfactual reasoning, and efficient learning. In this sense, world models provide a unifying mechanism for integrating prediction, planning, and generalization within sequential decision-making systems.

Despite rapid progress, the term world model is used inconsistently across the literature. Traditionally, it refers to an explicit model of environment dynamics that supports rollouts, planning, and counterfactual reasoning. More recently, the term has also been used in a broader sense to describe implicit internal representations that capture predictive regularities without exposing a standalone dynamics function. This conceptual ambiguity has blurred the boundaries between different approaches, making it difficult for newcomers to navigate the landscape and for practitioners to compare methods across domains. A clear and systematic tutorial is therefore needed to organize existing approaches, clarify terminology, and highlight the key design choices underlying different world modeling paradigms.

In this tutorial, we focus on explicit world models as a clear conceptual starting point and use them to build intuition about how world modeling supports prediction, planning, and learning. We explain what it means to model environment dynamics, why such models are useful, and how different design choices affect an agent’s behavior and capabilities. Implicit representations are introduced alongside explicit models, helping readers understand both their strengths and limitations. Rather than attempting to cover all existing methods exhaustively, this tutorial emphasizes core concepts, illustrative examples, and unifying principles. The material is organized to progressively guide the reader from basic ideas to more advanced topics, providing a structured pathway for understanding the diverse landscape of world models.

Beyond conceptual foundations, this tutorial also explores the role of world models in physical AI, where agents are embodied and interact with the physical world through perception and action. In such settings, world models are not merely abstract predictors but play a practical role in grounding perception, reasoning about dynamics, and coordinating long-horizon behaviors under physical constraints \cite{Fung2025}. We illustrate how world models enable embodied agents to anticipate the consequences of actions, perform imagination-based planning, and learn safely and efficiently when real-world interaction is expensive or risky. By connecting world modeling techniques to physical AI applications such as robotics and autonomous driving, this tutorial demonstrates how core ideas translate from algorithmic formulations to real-world intelligent behavior.

While a number of survey papers have reviewed specific aspects of world models or model-based reinforcement learning, most existing works primarily focus on cataloging methods and empirical results \cite{Ding2025, Kong2025, Zhu2024}. In contrast, tutorial-style treatments that systematically build intuition, clarify core concepts, and guide readers through fundamental design choices remain scarce. Moreover, to the best of our knowledge, no tutorial exists that jointly examines world models and their role in physical AI within a unified conceptual framework. By explicitly connecting world modeling principles to embodied settings such as robotics and autonomous driving, this tutorial fills an important gap in the literature. We believe that this integrated perspective—bridging conceptual foundations, algorithmic formulations, and physical-world applications—constitutes a key strength of the present tutorial.

The remainder of this paper is organized as follows. Section 2 introduces the fundamentals of world models. Section 3 reviews explicit world models for latent dynamics learning and planning. Section 4 discusses implicit world models based on large-scale representation learning. Section 5 extends these concepts to Physical AI in robotics and autonomous driving. Section 6 discusses pathways and challenges toward AGI. Finally, Section 7 concludes the paper.

\section{Fundamentals of World Models}
World models describe how the world evolves over time and how actions influence future outcomes. These ideas are closely related to human cognition, where the ability to predict and imagine future consequences plays a central role in intelligent behavior. Among the major paradigms of machine learning, reinforcement learning (RL) provides a natural framework for studying world models in artificial agents. In this section, we review the fundamentals of RL and use this framework to introduce the operational principles and design space of world models.

\subsection{World Models in Human Cognition: Motivation}

Figure 1 illustrates a familiar yet powerful example of how humans rely on a world model to reason beyond immediate perception \cite{Hawkins2021, Quiroga2005}. From a single snapshot of a diver mid-air, a human observer can infer unobserved aspects of the scene, such as the presence of a pool beneath the platform, even if it lies outside the current field of view. Based on prior experience, the observer can further predict how the scene will unfold over the next one or two seconds: the camera viewpoint will shift downward, the diver will enter the water, and a characteristic splash will follow. At the same time, the same world model supports counterfactual reasoning—if no water were present below, the observer immediately recognizes that the same action would lead to a catastrophic outcome. Moreover, subtle cues in the diver’s posture and body alignment allow an experienced observer to anticipate the quality of the performance and even predict that the dive is likely to receive a high score. These simultaneous inferences, predictions, and evaluations are not directly observable from the image itself, but instead arise from an internalized model that encodes physical dynamics, causal consequences, and task-specific regularities of the world.
\begin{figure}[h]
  \centering
  \includegraphics[width=0.25\linewidth]{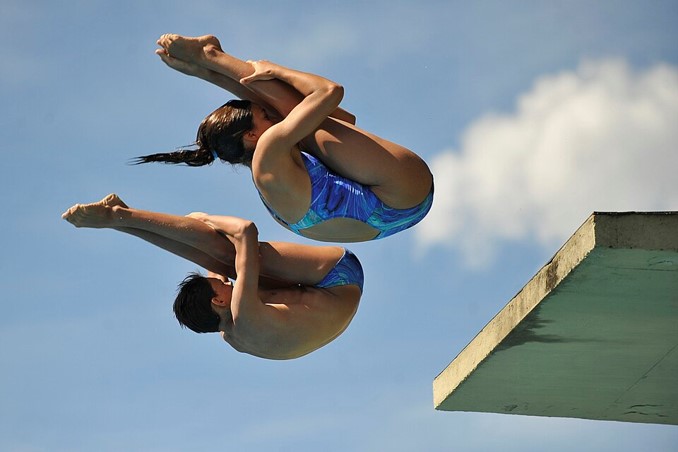}
  \caption{Multiple imaginative predictions inferred from a single partial observation through a world model (Source: Agência Brasil, CC BY 3.0).}
  \Description{world model}
\end{figure}

While Figure 1 provides an intuitive illustration of how humans reason beyond immediate perception, these capabilities can be more systematically understood through a hierarchical view of reasoning, as summarized in Figure 2 \cite{Pearl2009}. At the most basic level, humans exploit associative regularities by observing correlations in the world, enabling them to predict what is likely to happen next given a current situation. Beyond passive prediction, humans reason at the level of intervention, mentally simulating how the world would evolve if a particular action were taken. This allows them to evaluate alternative actions and anticipate their consequences before acting. At the highest level, humans engage in counterfactual reasoning, considering how outcomes would differ under hypothetical changes to the world or their own actions. Importantly, these levels are not isolated stages but coexist and interact within a unified internal model of the world. The hierarchical structure highlighted in Figure 2 thus provides a conceptual framework for understanding how world models support prediction, planning, and imagination in human cognition, and directly motivates the design of world models in artificial agents.
\begin{figure}[h]
  \centering
  \includegraphics[width=0.8\linewidth]{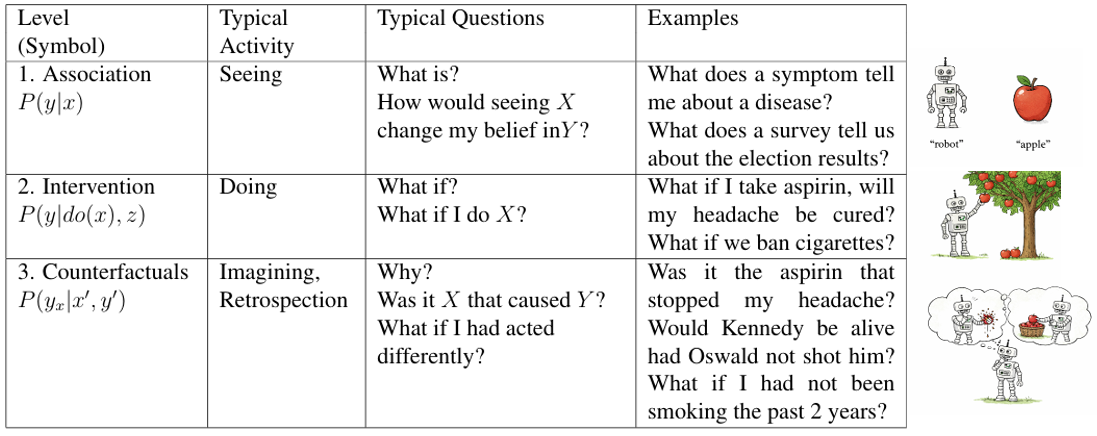}
  \caption{A conceptual illustration of hierarchical human reasoning supported by world models, inspired by the causal hierarchy of Judea Pearl (adapted from https://web.cs.ucla.edu/~kaoru/3-layer-causal-hierarchy.pdf).}
  \Description{Judea Pearl}
\end{figure}

\subsection{Reinforcement Learning and Sequential Decision Making}
RL formalizes sequential decision-making problems using the Markov decision process (MDP) framework \cite{Sutton2018}. An MDP provides a mathematical description of how an agent interacts with an environment over time and how its actions influence future states and rewards. This formulation serves as the foundational setting for defining and studying world models.

An MDP is defined by a tuple \((\mathcal{S}, \mathcal{A}, \mathscr{r}, \gamma)\) where \(\mathcal{S}\) denotes the state space, \(\mathcal{A}\) the action space, \(\mathscr{r}(s,a,s')\) the reward function, and \(\gamma \in [0,1]\) the discount factor. At each time step \(t\), the agent observes a state \(s_t \in \mathcal{S}\), selects an action \(a_t \in \mathcal{A}\) according to a policy \(\pi(a_t | s_t)\), and the environment transitions to a new state \(s_{t+1}\) while emitting a reward \(r_t\). Figure 3(a) illustrates this process. The state transition is governed by the environment dynamics, which define the probability distribution \(p(s_{t+1} |s_t,a_t)\).

\begin{figure}[h]
  \centering
  \includegraphics[width=0.6\linewidth]{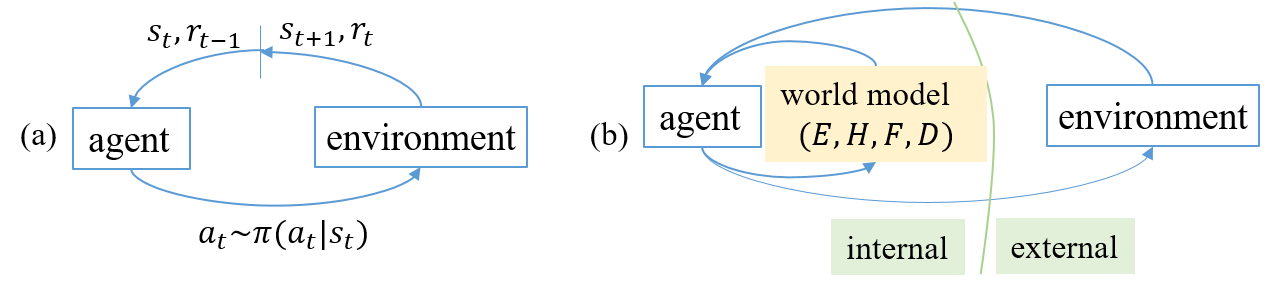}
  \caption{Comparison between (a) a standard MDP, in which the agent interacts directly with the external environment, and (b) an agent equipped with a world model, where environment dynamics are partially internalized through learned components \((E,H,F,D)\), enabling internal simulation and imagined rollouts.}
  \Description{MDP and world model}
\end{figure}

A key concept in RL is the trajectory, which represents a sequence of interactions between the agent and the environment: \(\tau=(s_0,a_0,r_0,s_1,a_1,r_1,\cdots,a_{T-1},r_{T-1},s_T)\)\footnote{Throughout this paper, trajectory refers to a sequence of transitions generated by interaction with the real environment, whereas rollout denotes a sequence of transitions imagined or simulated by a world model or planning procedure in latent space.}.  A trajectory is generated by repeatedly sampling actions from the policy and states from the environment dynamics until a terminal state is reached. Algorithm 1 illustrates the standard procedure for generating a trajectory in an MDP. Starting from an initial state sampled from the environment, the agent alternates between selecting actions using the current policy and receiving state transitions. This interaction loop constitutes the basic data-generation process in RL.

\begin{algorithm}
\small
\caption{Generating a trajectory from the environment}
\KwIn{MDP environment \textit{env} and policy network $\pi_\theta$}
\KwOut{a trajectory $\tau$}
state $\leftarrow$ \textit{env}.reset()\;
\While{True}{
    action $\sim \pi_\theta$(state)\;
    state1, reward, done $\leftarrow$ \textit{env}.step(action)\;
    state $\leftarrow$ state1\;
    \If{done}{
        break\;
    }
}
\end{algorithm}

The objective of RL is to find an optimal policy \(\pi_{\theta_*}\) that maximizes the expected cumulative reward over trajectories. The expected cumulative reward is commonly expressed as \(\mathbb{E}_{\tau \sim \pi_\theta}\left[\sum_{t=0}^{T-1} \gamma^t r_t\right]\)
 where \(T\) denotes the trajectory length. Algorithm 2 summarizes the generic structure of RL for policy optimization. The agent repeatedly generates trajectories using the current policy, evaluates performance using the collected data, and updates the policy parameters accordingly.

 \begin{algorithm}
 \small
\caption{Reinforcement learning for the optimal policy}
\KwIn{MDP environment \textit{env}}
\KwOut{optimal policy network $\pi_{\theta_*}$}
Initialize the parameters $\theta$ of the policy network $\pi_\theta$\;
\While{not converged}{
    Generate a trajectory $\tau$ using Algorithm 1\;
    Compute a loss from $\tau$\;
    Update $\theta$ using an optimizer\;
}
\end{algorithm}

This formulation makes explicit the central role of environment dynamics, trajectories, and policy optimization in RL. In the following sections, we will see how world models intervene in this process by learning an internal approximation of the environment dynamics and by enabling agents to generate trajectories without direct interaction with the real environment.

\subsection{Operational Principles of World Models}
While the trajectory-generation process described in Algorithm 1 is general, its most significant limitation is that the learning process is inherently tied to a specific task. When a new task is introduced, the MDP must be redefined with a new reward function and possibly a new state or action space, and the policy must be learned again from scratch. Knowledge acquired from previous tasks is not explicitly preserved or reused, making it difficult to support flexible adaptation, transfer, or compositional generalization. From this perspective, RL based solely on Algorithm 1 falls short of the notion of general intelligence, in which agents are expected to leverage prior knowledge to handle novel tasks efficiently. In addition, learning with Algorithm 2 alone relies on repeated interactions with the real environment to accumulate experience. This reliance directly impacts data efficiency, as large numbers of trajectories may be required before meaningful improvement is achieved. In many practical settings—such as robotics or autonomous driving—real-world interaction can be costly, slow, or unsafe, further limiting the applicability of purely interaction-driven learning.

World models address these limitations by decoupling the representation of environment dynamics from task-specific objectives. Instead of learning behavior solely through reward-driven interaction, an agent first learns a predictive model of how the environment evolves in response to actions. Once acquired, this model can be reused across different tasks by redefining goals or reward functions without relearning the underlying dynamics from scratch. In this sense, world models enable a form of knowledge accumulation in which experience from past interactions contributes to future learning and adaptation. This separation between learning how the world works and deciding what to do is a key step toward more general and flexible intelligence.

From an operational perspective, world models can be understood as internal simulators that allow an agent to mentally rehearse future interactions with the environment. Rather than directly querying the real world, the agent uses the world model to simulate possible futures and evaluate the consequences of its actions internally. This internal simulation capability typically relies on three key factors.

\begin{itemize}
\item {\textbf{Latent representation space}}: Many modern world models operate in a latent representation space, where high-dimensional observations are compressed into compact states that capture task-relevant structure while discarding irrelevant details. This latent space can naturally integrate multimodal observations—such as visual, auditory, and proprioceptive signals—and provide a shared representation across different tasks.
\item{\textbf{History-conditioned dynamics}}: The simulated dynamics are often conditioned on the history of observations and actions, enabling the model to infer unobserved or partially observable aspects of the environment and to maintain temporal coherence over long horizons.
\item{\textbf{Stochastic latent sampling}}: Many world models incorporate stochastic sampling of latent states, allowing them to represent uncertainty and generate diverse plausible futures rather than a single deterministic prediction.
\end{itemize}

Together, these factors enable world models to function as flexible internal simulators that support imagination, planning, and robust decision-making under uncertainty. To make these operational principles concrete, it is helpful to express them as a generic rollout procedure. A world model based on the rollout procedure can be described using four components: a recurrent state update function \(H\), an encoder \(E\), a decoder \(D\), and a latent dynamics model \(F\), as illustrated in Figure 3(b). Equation (1) defines their operations.\footnote{Here, \(=\) denotes deterministic computation, whereas \(\sim\) denotes stochastic sampling from a probability distribution, the distributions parameterized by \(q\) represent posterior inference conditioned on observations, while those parameterized by \(p\) represent prior or predictive distributions generated by the latent dynamics model.}

\begin{equation}
\left.
\begin{aligned}
\text{recurrent state update } H: \quad & h_t = H_{\phi_H}(h_{t-1}, z_{t-1}, a_{t-1}) \\
\text{encoder } E: \quad & z_t \sim q_{\phi_E}(z_t | h_t, x_t) \\
\text{latent dynamics } F: \quad & \hat{z}_t \sim p_{\phi_F}(\hat{z}_t | h_t) \\
\text{decoder } D: \quad & \hat{x}_t \sim p_{\phi_D}(\hat{x}_t | h_t, z_t)
\end{aligned}
\right\}
\end{equation}

While Equation (1) defines the components statically, Figure 4 illustrates their temporal interaction through a rollout procedure. To account for temporal dependencies and partial observability, the world model maintains a recurrent state \(h_t\), which aggregates information over time by \(H\). The recurrent state enables the model to retain information that may not be directly observable from a single observation, thereby supporting long-horizon reasoning and coherent state estimation. Given an observation \(x_t\) of multimodal high-dimensional sensory inputs, the encoder maps the inputs into a compact latent state representation by \(E\). This latent state provides an abstract representation that captures task-relevant structure while discarding irrelevant details. Such latent representations can unify multimodal observations and support representation sharing across tasks. The evolution of the latent state is governed by a dynamics model \(F\), where transitions are typically modeled as conditional probability distributions rather than deterministic mappings. This stochastic formulation allows the world model to represent uncertainty and to generate multiple plausible future trajectories through sampling, rather than committing to a single predicted outcome. The decoder \(D\) models the observation distribution, reconstructing observations from latent states for learning and interpretation.

The four components in Equation (1) play complementary roles in realizing a world model as an internal simulator. Algorithm 3 presents a canonical example of imaginary rollout generation using them as a world model, as illustrated in Figure 4. Starting from an observation \(x_k\) and a recurrent state \(h_k\) at time step \(k\), the world model maps observations into latent states and then repeatedly samples actions from the policy and state transitions from the learned dynamics model. While different world models instantiate these components in different ways, the algorithm captures a common operational structure shared by many modern approaches.

\begin{figure}[h]
  \centering
  \includegraphics[width=0.8\linewidth]{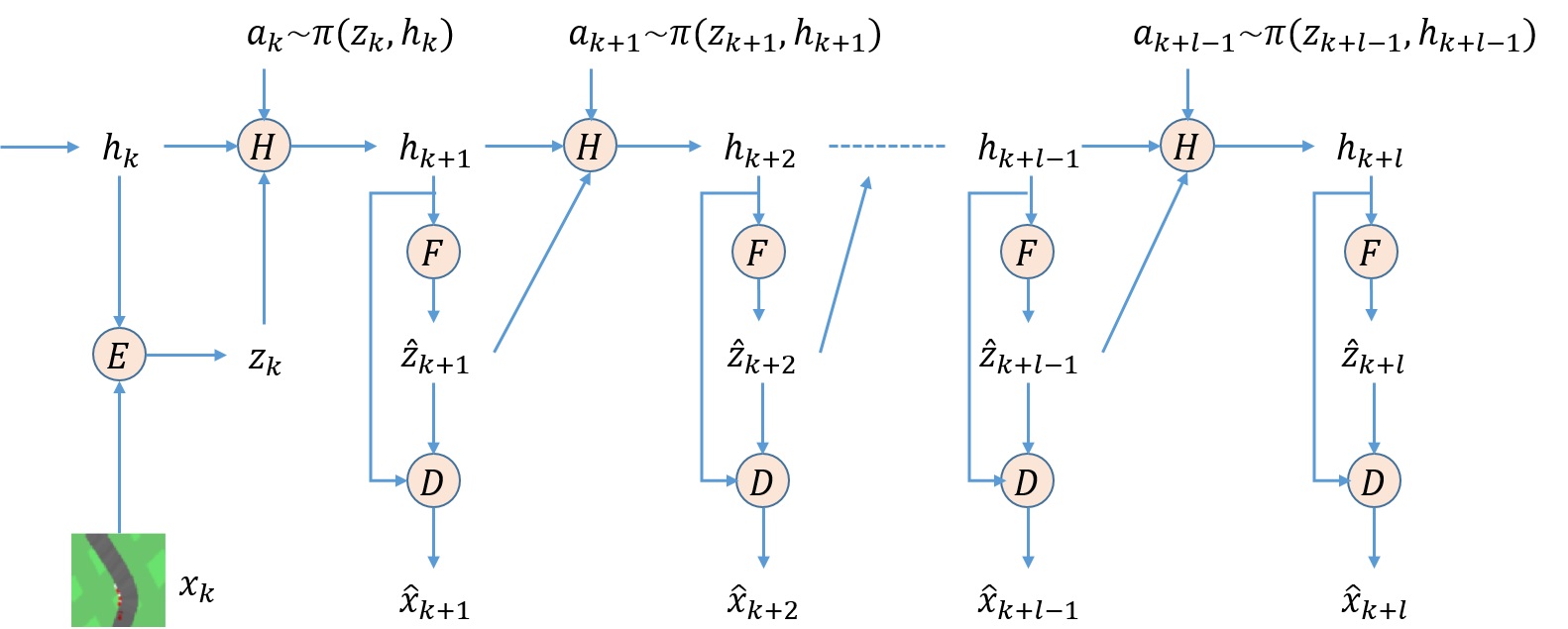}
  \caption{Imagined rollout generation using the world model in Algorithm 3. Starting from an encoded observation, the recurrent update model \(H\), latent dynamics model \(F\), policy \(\pi\), and decoder \(D\) iteratively generate future latent states, actions, and reconstructed observations in latent space.}
  \Description{rollout}
\end{figure}

\begin{algorithm}
\small
\caption{Generating an imaginary rollout using a world model}
\KwIn{world model $(E, H, F, D)$, policy $\pi$, time step $k$, history $h_{k}$, observation $x_k$}
\KwOut{rollout starting from time step $k$ with horizon $l$}
$z_k \sim q_{\phi_E}(\cdot | h_k, x_k)$ \tcp{transform observation into latent representation}
\For{$i = 0, 1, \cdots, l-1$}{
    $a_{k+i} \sim \pi(\cdot | h_{k+i}, z_{k+i})$ \tcp{action selection}
    $h_{k+i+1} = H_{\phi_H}(h_{k+i}, z_{k+i}, a_{k+i})$ \tcp{history update}
    $z_{k+i+1} \sim p_{\phi_F}(\cdot | h_{k+i+1})$ \tcp{transition to next time step}
    $\hat{x}_{k+i+1} \sim p_{\phi_D}(\cdot | h_{k+i+1}, z_{k+i+1})$ \tcp{reconstruction}
}
\end{algorithm}

Importantly, Algorithm 3 is not tied to a specific task or reward function. Once the world model has learned environment dynamics, the same model can be reused to generate imagined rollouts under different policies, objectives, or planning strategies. This flexibility enables agents to evaluate alternative behaviors, perform counterfactual reasoning, and adapt to new tasks without redefining the underlying dynamics or collecting new real-world data. In this sense, imaginary rollouts serve as a central mechanism through which world models support generalization, data efficiency, and long-horizon reasoning.

\subsection{Design Space of World Models}
It should be noted that Algorithm 3 represents a generic operational template rather than a concrete implementation. In practice, world models differ substantially in how these principles are instantiated—for example, in how latent states are defined, how histories are maintained, how stochastic transitions are parameterized, and how imagined rollouts are used for decision making. These variations give rise to a broad design space of world models. One useful way to understand this space is to consider several fundamental axes along which models differ, including the form of state representation (explicit state variables versus learned latent representations), the modeling of dynamics (deterministic or stochastic transitions and the treatment of uncertainty), the way the model is used (explicit simulation and planning through imagined rollouts versus predictive representations that shape policy learning), and the training objectives employed (such as reconstruction, reward prediction, or self-supervised learning signals).

Among these axes, a particularly consequential distinction concerns how explicitly environment dynamics are represented and used. In explicit world models, the agent learns a transition model that can be directly queried to simulate future states, rewards, or observations, enabling explicit rollouts, planning, and counterfactual reasoning. In contrast, implicit world models do not expose a standalone dynamics function; instead, predictive structure is embedded within learned representations or policies, influencing behavior without supporting direct simulation. Because this distinction has important implications for data efficiency, interpretability, and the forms of reasoning an agent can perform, we adopt the explicit–implicit distinction as a central organizing principle in the remainder of this tutorial. Section III begins by examining explicit world models that directly model environment dynamics and support simulation and planning.

\section{Explicit World Models}
Explicit world models enable an agent to reason about the future by explicitly parameterizing how the world evolves under actions. By providing direct access to a learned transition mechanism, these models support internal simulation and imagination, which, in turn, enable planning and counterfactual reasoning.

In this section, we develop a unified understanding of explicit world models by organizing these design choices around their underlying motivations and capabilities. Rather than presenting individual models in isolation, we focus on the common principles that govern explicit dynamics modeling, latent state learning, and imagination-based planning. Representative approaches—including the early world model of Ha and Schmidhuber \cite{Ha2018}, the Dreamer family of stochastic latent dynamics models \cite{Hafner2025}, and the planning-oriented MuZero architecture \cite{Schrittwieser2020}—are introduced as concrete instantiations of these principles. This perspective allows us to explain both the similarities and differences among explicit world models within a single coherent framework.

\subsection{Why Explicit Dynamics? Motivations for Generalizable Intelligence}
Section 2.3 introduced a generic operational template for world models, formalizing how the components \(E\), \(H\), \(F\), and \(D\) interact to generate imaginary rollouts through Algorithm 3. While this formulation specifies how a world model operates, it does not yet explain why such a decomposition is advantageous for intelligent systems that must generalize across tasks. This subsection motivates explicit dynamics modeling by examining how it supports abstraction, task generalization, and the reuse of knowledge. 

\textbf{Encoding observations into a latent space ($E$):} Intelligent agents operate on high-dimensional, heterogeneous, and often noisy observations. Directly learning task-specific policies in such spaces typically yields representations tightly coupled to a particular objective, limiting transfer to new tasks. Mapping observations into a latent space separates representation from task objectives, allowing the encoder $E$ to capture predictive structure rather than reward-specific features. The resulting latent space serves as a task-agnostic abstraction of the environment, enabling reuse across different tasks without relearning perception.

\textbf{Maintaining a history-dependent state ($H$):} In partially observable environments, relevant state variables are not directly accessible at each time step. The recurrent state update function $H$ addresses this by aggregating information over time, forming a state that reflects the underlying environment rather than the immediate observation. This separation prevents task-specific heuristics from being embedded in the policy and allows the same latent representation to support multiple tasks, including prediction, control, and planning.

\textbf{Modeling dynamics explicitly in latent space ($F$):} A key limitation of standard RL is the entanglement of environment dynamics with task-specific rewards. Explicit latent dynamics modeling resolves this by decoupling what the world does from what the agent wants. The transition model $F$ captures how latent states evolve under actions, independent of any particular objective. As a result, learned dynamics can be reused across tasks, with new objectives defined through goals, costs, or planning strategies. This separation enables adaptation to new tasks without relearning environment behavior from scratch.

\textbf{Reconstructing observations ($D$):} Reconstruction grounds latent dynamics in observable experience by constraining internal states to remain predictive of sensory inputs. This reduces the risk of drifting toward task-specific or spurious abstractions and improves robustness across tasks. However, reconstruction is not always necessary: in some settings, latent states need only support planning or decision making. This highlights a trade-off between perceptual grounding and abstraction, making the inclusion of $D$ a design choice rather than a requirement.

Taken together, these components show that explicit world models are not merely tools for prediction or planning, but mechanisms for organizing knowledge in a task-independent manner. By separating representation, history-dependent state maintenance, dynamics, and decision making, they enable agents to learn how the world evolves and reuse this knowledge across tasks with minimal additional learning. This decomposition supports task generalization, long-horizon reasoning, and adaptation to new objectives. This perspective parallels hierarchical human reasoning. As illustrated in Figure 2 and inspired by Pearl’s causal hierarchy \cite{Pearl2009}, reasoning progresses from association to intervention and ultimately to counterfactual reasoning. By explicitly modeling environment dynamics, world models enable agents not only to predict future observations but also to evaluate interventions and reason about alternative trajectories, providing a computational pathway toward more general forms of intelligence.

\subsection{Learning Latent World Models: A Common Training Paradigm}
Although explicit world models differ in architecture and application, many representative approaches—including early latent world models, Dreamer-style stochastic models, and planning-oriented architectures such as MuZero—share a common learning paradigm. This subsection abstracts away model-specific details and highlights the core training principles that underlie these approaches.

\textbf{Neural parameterization:} In modern world model-based systems, the components introduced in Section 2.3—encoder $E$, recurrent state update function $H$, dynamics model $F$, and decoder $D$—are typically implemented as neural networks with parameters $\phi = \{\phi_E, \phi_H, \phi_F, \phi_D\}$, while the policy $\pi$ is parameterized separately by $\theta$. This parameterization enables end-to-end learning from data using gradient-based optimization and provides the flexibility required to model high-dimensional observations and complex dynamics. More importantly, it establishes a common interface that separates representation learning and dynamics modeling from decision making.

\textbf{Training signals:} Learning a world model requires aligning model predictions with real experience. Training primarily relies on trajectories collected from the environment, denoted by $\mathcal{D}^{\text{real}}$, consisting of sequences of observations and actions. At the same time, the model generates imagined rollouts by unrolling latent dynamics, yielding a distribution of rollouts $\mathcal{D}^{\text{latent}}$. Learning is thus formulated as minimizing discrepancies between quantities inferred from $\mathcal{D}^{\text{real}}$ and predictions derived from $\mathcal{D}^{\text{latent}}$, defined in observation space, latent space, or both. This leads to a key design choice: how information from real experience constrains imagined rollouts. Reconstruction losses, latent consistency objectives, and reward or value prediction errors all serve to align $\mathcal{D}^{\text{latent}}$ with the statistics of $\mathcal{D}^{\text{real}}$. Regardless of the specific formulation, the central objective is to ensure that imagined rollouts remain consistent with the real environment.

\textbf{Optimization strategies:} Another fundamental design decision concerns how the world model and policy are trained. In joint training, the parameters $\phi$ and $\theta$ are optimized simultaneously using a combined objective, allowing representation, dynamics, and control to co-adapt. Alternatively, training can proceed in a staged or alternating manner: the world model is first learned from real data, after which the policy is optimized using imagined rollouts, followed by further model updates under the improved policy. Joint training can yield compact, behavior-adapted representations, whereas staged training often improves stability and interpretability.

Explicit world models occupy different points along the design spectrum, reflecting trade-offs among stability, sample efficiency, and generalization. Importantly, these choices do not alter the underlying operational template described in Section 2.3, but rather determine how effectively it is realized in practice. The following subsections examine how this common structure is instantiated in concrete models, progressing from early latent world models to more expressive and planning-oriented designs.

\subsection{Memoryless Latent World Models: The Ha–Schmidhuber Approach}
The operational framework introduced in Section 2.3 can be instantiated in its most transparent and pedagogically useful form by early latent world models, most notably the approach proposed by Ha and Schmidhuber \cite{Ha2018}. Although limited in scale compared to modern systems, this model serves as a representative example of how perception, dynamics modeling, and control can be explicitly separated, trained, and recombined into a functioning world model-based agent.

\textbf{Architectural decomposition:} The Ha–Schmidhuber model realizes the components $E$, $H$, $F$, and $D$ through a modular neural architecture, following a staged decomposition in which perception, temporal modeling, and control are implemented as distinct modules. Figure 5(a) presents a single-step view of the rollout process in Figure 4, explicitly showing the inputs and outputs of these components at a given time step $t$. 

Perception is implemented using a variational autoencoder, where the encoder $E$ maps observations $x_t$ to a stochastic latent variable $z_t$, and the decoder $D$ reconstructs observations. Crucially, the encoding is independent of temporal context, reducing the general form in Equation (1) to $z_t=q(z_t |x_t)$, yielding a memoryless latent representation that does not incorporate historical information.

Temporal dynamics are modeled by a recurrent neural network with a mixture density network output (MDN-RNN). The recurrent hidden state $h_t$ evolves deterministically to summarize past information, while the dynamics model $F$ is implemented as a MDN that models the next latent state as a Gaussian mixture distribution. This stochastic parameterization enables the model to capture uncertainty and generate multiple plausible future trajectories, while restricting the role of memory to prediction rather than state inference.

\begin{figure}[h]
  \centering
  \includegraphics[width=0.7\linewidth]{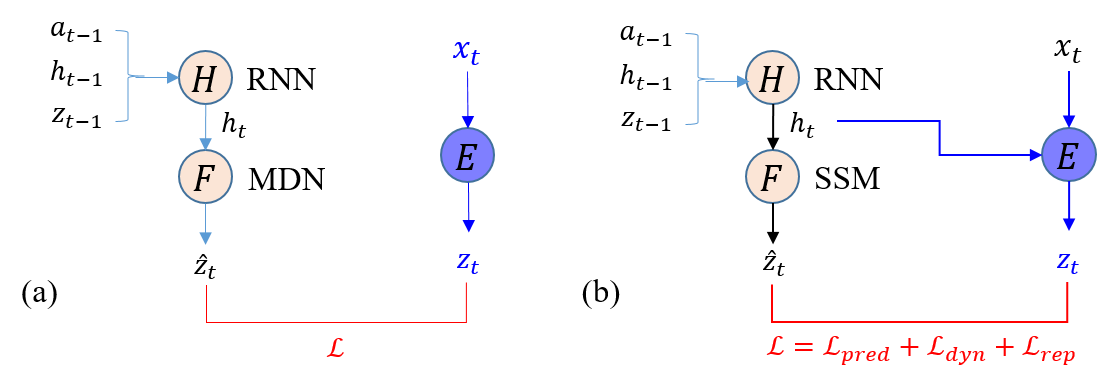}
  \caption{Architectural comparison of latent world models. (a) Ha–Schmidhuber model (RNN-MDN): the encoder is independent of the recurrent state, resulting in memoryless latent inference. (b) Dreamer model (RSSM): the encoder is conditioned on the recurrent state, enabling history-conditioned inference.}
  \Description{architecture comparison}
\end{figure}

The policy \(\pi\) is implemented as a simple controller that takes both the recurrent state \(h_t\) and the current latent \(z_t\) as input. This design reflects a key principle: expressive capacity is concentrated in the world model, while the policy exploits the predictive structure encoded in latent dynamics to select actions. By relying on the learned internal model, the policy focuses on decision making rather than compensating for representational limitations.

In the original terminology of Ha and Schmidhuber \cite{Ha2018}, the system is decomposed into a vision model (V), a memory model (M), and a controller (C), corresponding to \((E,D)\), \((H,F)\), and \(\pi\), respectively. It is important to note that, while environment states are represented in latent space, actions and rewards remain grounded in the real interaction space. The world model internalizes state dynamics, but control and evaluation are defined with respect to the external environment.

\textbf{Training procedure:} The Ha–Schmidhuber model is trained in a staged manner, as illustrated in Figure 6, where representation learning, dynamics modeling, and policy optimization are performed sequentially rather than end-to-end. This decomposition stabilizes learning by first establishing a latent representation before modeling temporal dynamics.

\begin{figure}[h]
  \centering
  \includegraphics[width=0.9\linewidth]{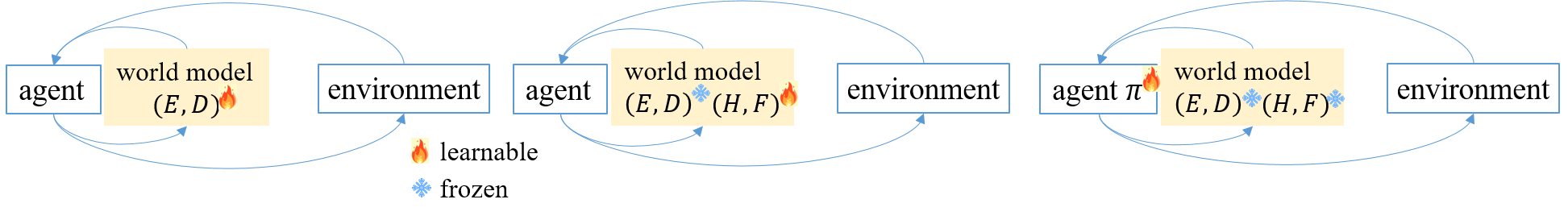}
  \caption{Staged learning procedure of the Ha–Schmidhuber world model. The encoder–decoder pair \((E,D)\), latent dynamics components \((H,F)\), and policy \(\pi\) are trained sequentially, with previously learned components frozen at each stage.}
  \Description{architecture comparison}
\end{figure}

In the first stage, a variational autoencoder is trained on observations collected from the environment, producing an encoder that maps high-dimensional inputs to a compact latent space. In the second stage, the dynamics model is trained on sequences of latent variables and actions, learning to predict the distribution of the next latent state using a mixture density network, by minimizing the negative log-likelihood loss $\mathcal{L}$ of the encoded target under the predicted distribution (Figure 5(a)). The encoder remains fixed, and its outputs serve as targets for the dynamics predictor. In the final stage, the policy is optimized using the learned world model. Since the model can simulate environment dynamics in latent space, the controller can be trained using imagined rollouts, reducing the need for direct interaction with the real environment.

While this staged training procedure provides conceptual clarity and stability, it limits the model's ability to adapt its representations to task-specific objectives, motivating the development of end-to-end approaches such as Dreamer.

\textbf{Inference: real-world interaction vs. latent imagination:} The operational behavior of the Ha–Schmidhuber model can be understood by contrasting two modes of inference, corresponding to the external and internal interaction loops in Figure 3(b): real-world interaction and latent imagination.

During real-world interaction, the agent processes incoming observations through the encoder \(E\) to produce the latent state \(z_t\), which is combined with the recurrent state \(h_t\) to update the internal state and select actions. In this mode, the model remains grounded in external observations, ensuring that the latent representation reflects the current state of the environment.

In contrast, during latent imagination, the model operates entirely within its learned dynamics. Instead of receiving observations from the environment, the dynamics model \(F\) generates predicted latent states, which are recursively fed back into the system, as illustrated in Figure 4 and Algorithm 3. This creates a self-contained simulation in latent space, allowing the agent to explore possible future rollouts without interacting with the real environment. By unrolling this internal simulation, the world model functions as an internal playground, enabling anticipation of future outcomes and policy refinement without real-world interaction.

This dual-mode operation highlights the world model's role as an internal simulator, enabling decision-making through imagined rollouts while maintaining grounding in real-world observations.

\textbf{Implications and limitations:} The Ha–Schmidhuber model represents an early milestone in world model-based RL, demonstrating that complex behaviors can be learned by decomposing an agent into a predictive world model and a simple controller. Its central contribution is the introduction of latent imagination: by learning environment dynamics in latent space, the agent can simulate future rollouts and refine its policy without extensive real-world interaction.

Despite this conceptual clarity, the architecture exhibits several fundamental limitations. First, latent states are inferred solely from the current observation, resulting in memoryless inference that ignores temporal context. Moreover, although the model maintains a recurrent hidden state, it is deterministic and not used in the inference process. As a result, the model cannot represent uncertainty in the latent state or maintain multiple hypotheses over time, limiting its ability to handle partial observability. Second, the staged training procedure decouples representation learning, dynamics modeling, and control, restricting the model’s ability to adapt its internal representations to task-specific objectives.

These limitations highlight the need for models that integrate temporal context into state inference, represent uncertainty explicitly, and support end-to-end learning. This motivates the development of modern world models such as Dreamer, which address these challenges through stochastic latent states and history-conditioned inference.

\subsection{Contextual Latent World Models: The Dreamer Framework}
To address the limitations of memoryless inference and deterministic recurrent states in earlier models, the Dreamer framework introduces a contextual latent world model based on the Recurrent State-Space Model (RSSM) \cite{Hafner2025}. By conditioning latent state inference on both the current observation and the recurrent state, Dreamer enables uncertainty-aware latent state estimation under partial observability.

Building on this foundation, Dreamer integrates latent imagination with end-to-end learning, enabling agents to jointly learn dynamics, value functions, and policies within a unified framework. In this subsection, we examine its architectural design, training procedure, and inference mechanism, highlighting how these components support data-efficient and long-horizon decision making.

\textbf{Architectural decomposition: latent state modeling by RSSM:} At the core of the Dreamer framework is the RSSM, illustrated in Figure 7, which combines recurrent neural networks (RNNs, typically implemented as GRUs or LSTMs) and stochastic state-space models (SSMs, e.g., Gaussian or categorical latent transitions) to represent the system state as a deterministic recurrent state \(h_t\) and a stochastic latent variable \(z_t\). This hybrid structure captures both long-term temporal dependencies and uncertainty within a unified latent state representation.

\begin{figure}[h]
  \centering
  \includegraphics[width=0.8\linewidth]{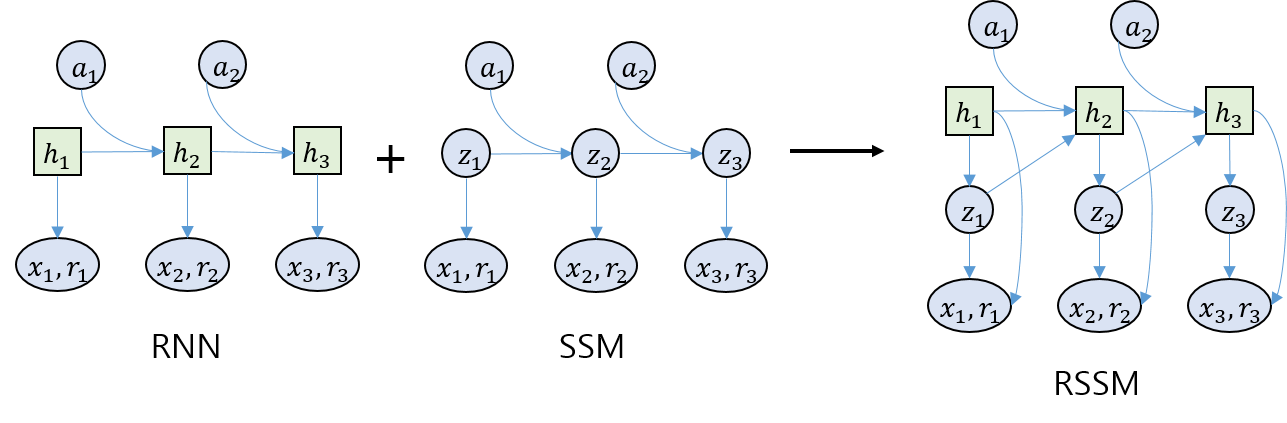}
  \caption{Recurrent State-Space Model (RSSM). RSSM represents the system state using a deterministic recurrent state \(h_t\) and a stochastic latent variable \(z_t\), combining temporal structure and uncertainty in a unified latent state representation.}
  \Description{RSSM illustration}
\end{figure}

A key distinction from earlier world models lies in how latent states are inferred. As shown in Figure 5(b), Dreamer conditions the encoder on both the observation and the recurrent state, i.e., \(z_t\sim q(z_t |h_t,x_t)\), enabling history-conditioned latent inference under partial observability, in contrast to the memoryless inference used in earlier models.

The deterministic component \(h_t\) captures long-term temporal structure, while the stochastic component \(z_t\) represents uncertainty and multimodal dynamics. By combining these complementary roles, RSSM overcomes the limitations of memoryless inference and supports robust state estimation over time.

Building on this latent representation, Dreamer extends the generic world model by incorporating reward and continuation models. In addition to reconstructing observations, the model predicts rewards and continuation signals directly from the latent state using Equation (5).\footnote{In neural networks implementing the distributions in Eq. (1) and Eq. (2), denoted by the sampling operator \(\sim\), the network predicts the parameters of the distribution, and sampling is performed using differentiable estimators (e.g., the reparameterization trick for continuous latents). The Dreamer instead employ discrete categorical latents trained with straight-through gradient estimation.} These additional prediction heads enable the model to evaluate imagined rollouts entirely in latent space, without requiring further interaction with the environment.

\begin{equation}
\left.
\begin{array}{ll}
\text{reward } R\text{:} & \hat{r}_t \sim p_{\phi_R}(\hat{r}_t | h_t, z_t) \\
\text{continuation } C\text{:} & \hat{c}_t \sim p_{\phi_C}(\hat{c}_t | h_t, z_t)
\end{array}
\right\}
\end{equation}

\textbf{Training procedure: joint optimization and KL alignment:} The Dreamer framework is trained end-to-end by jointly optimizing the world model, actor, and critic, forming a closed learning loop between real experience and latent imagination. As outlined in Algorithm 4, real trajectories are first used to update the world model, which then generates imagined rollouts for policy and value learning. Unlike earlier staged approaches, all components are optimized simultaneously, allowing representations, dynamics, and policies to co-adapt.

\begin{algorithm}
\small
\caption{Concurrent learning of world model and actor-critic}
\KwIn{MDP environment $env$}
\KwOut{trained policy network $\pi_{\theta_*}$ and world model $\phi$}
Initialize replay buffer $\mathcal{D}$\;
Initialize the parameters $\phi, \theta, \psi$ of the world model, actor, and critic, respectively\;
\While{not converged}{
    Collect a real trajectory $\tau$ by $\pi_\theta$ with latent states inferred from real observations\;
    Add $\tau$ to the replay buffer $\mathcal{D}$ and sample minibatch $B$ from $\mathcal{D}$\;
    Compute loss $\mathcal{L}(\phi)$ and gradients from $B$, and update $\phi$ \tcp{training world model}
    Generate an imaginary rollout $\tilde{\tau}$ using Algorithm 3\;
    Compute loss $\mathcal{L}(\psi)$ and gradients from $\tilde{\tau}$, and update $\psi$ \tcp{training critic}
    Compute loss $\mathcal{L}(\theta)$ and gradients from $\tilde{\tau}$, and update $\theta$ \tcp{training actor}
}
\end{algorithm}

The world model is trained to align imagination with reality by learning from real trajectories stored in the replay buffer. At its core is a variational objective that maintains two distributions over latent states: a posterior \(q_{\phi_E}(z_t | h_t, x_t)\), inferred from observations, and a prior \(p_{\phi_F}(\hat{z}_t |h_t)\), predicted from the recurrent state. The training objective enforces alignment between these distributions through a KL divergence loss, encouraging the prior to approximate the posterior. This alignment enables the model to generate realistic rollouts even in the absence of observations.

This objective can be implemented through a decomposition of the world model loss \(\mathcal{L}(\phi)\) into three components, as illustrated in Figure 5(b). The prediction loss \(\mathcal{L}_{pred}\) reconstructs observations, rewards, and continuation signals from latent states, ensuring that the representation remains grounded in real data. The dynamics loss \(\mathcal{L}_{dyn}\) trains the prior to match the posterior, enabling accurate prediction of latent transitions. The representation loss \(\mathcal{L}_{rep}\) regularizes the posterior toward the prior, preventing divergence between inference and prediction. Together, these components realize the KL alignment objective while maintaining a stable and informative latent representation.

Once the world model is trained, it supports imagination-based actor–critic learning. By unrolling latent dynamics from the learned prior, the agent generates imagined rollouts, from which the critic is trained to estimate long-term value and the actor is optimized to maximize expected returns. As both are defined over the latent state \(s_t=[h_t,z_t]\) (Equation (3)), policy improvement can occur entirely in latent space, significantly reducing reliance on real-world interaction. The same policy \(\pi_\theta\) is used for both real interaction and imagined rollouts, ensuring consistency between data collection and policy optimization, although the underlying state distributions differ.

\begin{equation}
a_t \sim \pi_\theta(a_t | s_t) \quad \text{and} \quad v_\psi(s_t)
\end{equation}

This joint optimization creates a synergy between the world model and the actor–critic. The world model learns a task-agnostic representation of environment dynamics, while the actor–critic adapts this representation for task-specific objectives, enabling data-efficient learning through extensive practice in latent imagination.

\textbf{Inference: latent state estimation and policy deployment:} In the Dreamer framework, inference operates at two complementary levels: external inference for interaction with the environment and internal inference for latent imagination. Together, these processes enable the agent to maintain a coherent latent state estimate over time while leveraging its learned world model for decision making.

During interaction with the environment, the agent performs external inference by updating its latent state using observations. As illustrated in Figure 5(b), at each time step, the encoder \(E\) processes the current observation \(x_t\) together with the recurrent state \(h_t\) to infer the posterior latent state \(z_t \sim q_{\phi_E}(z_t |h_t,x_t)\). This posterior inference grounds the latent representation in real observations, allowing the agent to maintain an accurate latent state estimate under partial observability. The actor \(\pi_\theta\) then maps the latent state \([h_t,z_t]\) to an action, and the recurrent state is updated via \(h_{t+1}=H_{\phi_H}(h_t,z_t,a_t)\), forming the standard perception–action loop.

In contrast, internal inference corresponds to imagination within the learned world model. As illustrated in Figure 4 and Algorithm 3, the model generates imagined rollouts by unrolling latent dynamics starting from a posterior-inferred state. After initialization, future latent states are predicted using the prior \(p_{\phi_F}(\hat{z}_t|h_t)\) without access to further observations. This process enables the agent to simulate possible futures and anticipate outcomes entirely in latent space.

The same policy \(\pi_\theta\) is applied in both external interaction and internal imagination, providing a unified decision mechanism across real and simulated trajectories. This enables seamless transitions between observation-grounded inference and model-based prediction.

\textbf{Implications and limitations:} The Dreamer framework demonstrates how explicit world models can enable data-efficient and long-horizon learning through latent imagination. By learning environment dynamics in a task-agnostic latent space and optimizing behavior through imagined rollouts, the agent can substantially reduce reliance on real-world interaction while maintaining coherent decision making over extended horizons. Empirically, a single model configuration has been shown to scale across a large number of diverse tasks—on the order of hundreds—highlighting the practical benefits of learning reusable latent dynamics.

Despite these strengths, the approach has inherent limitations. Because policy learning depends on the accuracy of the learned world model, modeling errors can accumulate over imagined rollouts and bias decision making, a phenomenon often referred to as model bias or model exploitation. Furthermore, while observations, rewards, and continuation signals are successfully abstracted into latent space, actions remain tightly coupled to specific tasks or embodiments, limiting transferability across different environments or control interfaces.

While Dreamer focuses on improving policies through imagined rollouts, alternative approaches leverage learned world models for planning at decision time. MuZero exemplifies this complementary paradigm by integrating latent dynamics with search.

\subsection{Planning-Oriented Latent World Models: The MuZero Framework}
Evolving from the AlphaGo line of research through AlphaGo Zero and AlphaZero \cite{Silver2018}, MuZero extends search-based decision making beyond board games to more general domains, including Atari environments \cite{Schrittwieser2020}. By learning environment dynamics directly from interaction, it integrates a latent world model with Monte Carlo Tree Search (MCTS), enabling planning in domains where the underlying rules are unknown.

\textbf{Architecture: latent dynamics and planning:} Figure 8(a) illustrates the architecture of MuZero, which integrates a learned latent world model with Monte Carlo Tree Search (MCTS) for decision-time planning. The model is defined by three functions—a representation function \(h\), a dynamics function \(g\), and a prediction function \(f\)—whose roles are summarized in Equation (3). These functions are jointly parameterized as a single neural network \(\mu_\theta\), and are trained end-to-end. Here, \(h\) and \(g\) play the roles of the encoder \(E\) and the latent dynamics \(F\) in Equation (1), respectively.

\begin{equation}
s^0 = h_\theta(x_{1:t});\quad (r^k, s^k) = g_\theta(s^{k-1}, a^k);\quad (p^k, v^k) = f_\theta(s^k)
\end{equation}

\begin{figure}[h]
  \centering
  \includegraphics[width=0.8\linewidth]{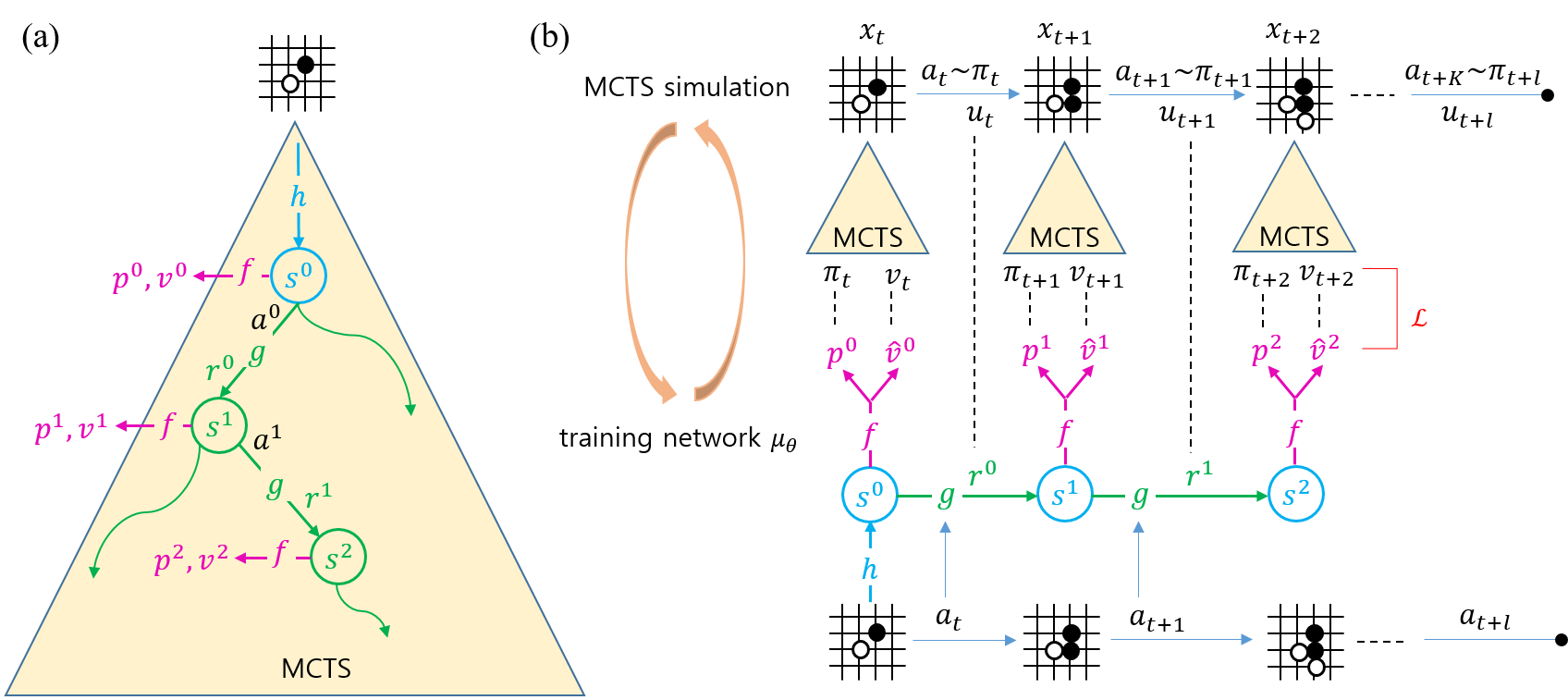}
  \caption{Planning-oriented world model learning in MuZero. (a) Monte Carlo Tree Search performed in latent space using the representation, dynamics, and prediction functions. (b) End-to-end learning with MCTS, where search-derived policy and value targets guide the optimization of the world model.}
  \Description{MuZero illustration}
\end{figure}

The representation function \(h\) maps the observation history \(x_{1:t}\) to an initial latent state \(s^0\) that summarizes the information required for decision making. Although conceptually defined over the full history, in practice a fixed-length sequence of recent observations (e.g., stacked frames) is used as input. This latent state serves as the root of the search tree. Starting from this state, the dynamics function \(g\) predicts the next latent state and the corresponding reward given an action, defining a learned transition model in latent space. Unlike AlphaZero, which has access to the true environment dynamics, MuZero learns this latent dynamics function directly from interaction data. The model is not required to reconstruct observations, but instead focuses on predicting quantities relevant for planning. The prediction function \(f\) outputs a policy prior \(p\) and a value estimate \(v\) for each latent state. These outputs guide the search process by providing both a prior over promising actions and an estimate of long-term return.

Planning is performed by MCTS in latent space, as illustrated in Figure 8(a), by unrolling the learned dynamics to simulate future states. Through this architecture, MuZero combines representation learning with decision-time planning, enabling action selection through search over learned latent dynamics rather than direct interaction with a known simulator.

\textbf{Training: learning from search-guided targets:} Figure 8(b) illustrates the training process of MuZero, where planning and representation learning are tightly integrated. Rather than training a policy directly from environment trajectories, MuZero uses Monte Carlo Tree Search (MCTS) to produce improved policy targets that guide learning.

Algorithm 5 summarizes the overall training loop. During interaction with the environment, trajectories are generated by repeatedly running MCTS and executing the selected actions. At each step, MCTS uses the current networks \(\mu_\theta\) to compute a search distribution \(\pi_t\), from which an action is sampled and applied to the environment, yielding the next observation and reward. The resulting trajectory \(\tau\) is stored in a replay buffer.

\begin{algorithm}
\small
\caption{MuZero training with search-guided planning}
\KwIn{environment $env$}
\KwOut{trained world model $\mu_\theta$}
Initialize replay buffer $\mathcal{D}$\;
Initialize parameters $\theta$ of $\mu_\theta$\;
\While{not converged}{
    Generate a trajectory $\tau$ by repeating; \tcp{planning process}
    \Indp
        Run MCTS using the current networks \(\mu_\theta\) to obtain search distribution $\pi_t$\;
        Select action $a_t \sim \pi_t$ and execute it in the environment, and observe $r_t, x_{t+1}$\;
    \Indm
    Store $\tau$ in $\mathcal{D}$\;
    Sample a trajectory segment $(x_i, a_i, u_i, \pi_i)_{i=t}^{t+l}$ from $\mathcal{D}$\;
    Unroll the latent dynamics model for $l$ steps starting from $s^0 = h(x_{1:t})$\;
    Compute the combined policy, value, and reward loss\;
    Update parameters $\theta$\;
}
\end{algorithm}

Training proceeds by sampling a trajectory segment (or minibatch of trajectory segments) from the replay buffer. Starting from the latent state \(s^0\), the dynamics function \(g\) is unrolled for \(l\) steps to generate a sequence of latent states. This unrolling allows gradients to propagate through the representation, dynamics, and prediction functions.

The networks are optimized using a combined loss consisting of policy, value, and reward terms. The policy loss aligns the predicted policy \(p^k\) with the search distribution \(\pi_{t+k}\). The value loss aligns the predicted value \(\hat{v}^k\) with the target return \(v_{t+k}\), computed from stored rewards and a bootstrap value predicted by the current network. The reward loss aligns the predicted reward \(r^k\) with the observed environment reward \(u_{t+k}\). Through these objectives, the model learns latent representations and transitions that support effective planning.

\textbf{Inference: decision-time planning:} During inference, the trained networks \(\mu_\theta\) are kept fixed and used for decision-time planning. At each step, the current observation is mapped to a latent state via the representation function \(h\), and MCTS is executed in latent space by repeatedly applying the learned dynamics model to simulate future states. The search produces a distribution over actions based on visit statistics, from which an action—typically the most visited—is selected and executed in the environment. This process corresponds to executing the planning process of Algorithm 5 without data storage or parameter updates, and is repeated at each decision step.

\textbf{Implications and Limitations:} MuZero demonstrates a powerful paradigm for planning-oriented world models by integrating representation learning, latent dynamics modeling, and search-based decision making. Unlike earlier AlphaGo-style systems that rely on known environment rules, MuZero learns a dynamics model directly from interaction and performs planning entirely in a learned latent space. A key implication of this design is the extension of planning-based RL beyond symbolic domains. While earlier search-based systems such as AlphaGo and AlphaZero were largely confined to environments with explicitly defined rules, MuZero enables planning in perceptual settings by learning environment dynamics in latent space. Empirically, it achieves strong performance in both board games and Atari, demonstrating that planning with learned dynamics can bridge symbolic domains and high-dimensional sensory environments.

Despite these advantages, planning-oriented world models introduce practical limitations. Decision-time planning requires repeated MCTS simulations, leading to higher computational cost than amortized policy methods. In addition, performance depends on the accuracy of the learned latent dynamics, and modeling errors can bias search outcomes over long horizons. Finally, because MuZero relies on tree search, it is naturally suited to discrete action spaces, and extending the framework to continuous control remains challenging.

\section{Implicit World Models}
Explicit world models capture the world by modeling environment dynamics and simulating future rollouts. Implicit world models, by contrast, dispense with explicit dynamics and encode world knowledge directly in learned representations. This shift reframes world modeling from simulation to representation, where reasoning emerges through inference over structured representations rather than explicit rollouts. The remainder of this section examines the implications of this perspective.

\subsection{Concept: World Knowledge as Representation}
Implicit world models embed world knowledge directly in learned representations. From this perspective, world modeling is not primarily about simulating future trajectories, but about organizing structural regularities of the environment within a representation space. Such regularities—e.g., object persistence, physical constraints, and action feasibility—can guide behavior even without explicit forward simulation.

In this view, the representation space functions as a structured coordinate system of the world. Similar situations are mapped to nearby representations, incompatible states are separated, and salient factors of variation align along meaningful dimensions. World knowledge is thus encoded in the geometry of representations rather than in an explicit transition model. Recent empirical evidence supports this perspective. Large pretrained models exhibit internal representations that capture spatial and temporal structure despite the absence of explicit world modeling supervision \cite{Gurnee2024, Feng2025a}. These findings suggest that meaningful world knowledge can emerge from representation structure alone.

The distinction between explicit and implicit world models is illustrated in Figure 9. While explicit models rely on internal dynamics simulators for rollout-based prediction, implicit models encode knowledge in representations and support reasoning through inference over structured latent spaces. Both paradigms can be interpreted through a two-loop perspective: an external loop for interaction with the environment and an internal loop for decision support. In explicit models, the internal loop performs simulation, whereas in implicit models it operates through inference over representations.

\begin{figure}[h]
  \centering
  \includegraphics[width=0.9\linewidth]{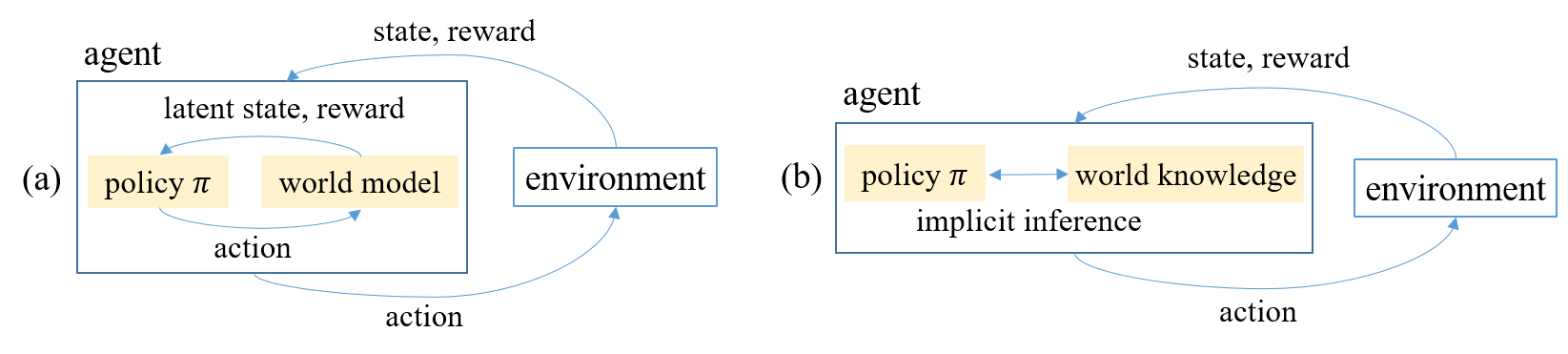}
  \caption{Explicit versus implicit world models. (a) Explicit world models encode world knowledge in an internal dynamics simulator that generates imagined rollouts. (b) Implicit world models embed world knowledge directly in learned representations, where inference over structured representations replaces explicit simulation as the primary mechanism for reasoning.}
  \Description{explicit vs. implicit}
\end{figure}

\subsection{Learning Paradigms for Implicit World Models}
Implicit world models are learned through large-scale predictive and self-supervised learning. Instead of fitting a transition function that simulates environment evolution, these models learn representations that encode regularities of the world by enforcing predictive consistency across observations.

The central mechanism underlying this paradigm is predictive learning. By requiring the model to predict missing, future, or alternative views of data, training implicitly pressures representations to capture a stable structure that makes observations coherent. Although predictive objectives vary—from next-token prediction in language models to masked modeling in multimodal systems and future-frame prediction in video—they share a common principle: accurate prediction requires sensitivity to persistent regularities such as object continuity, spatial relationships, and causal dependencies.

A useful way to understand this process is through the notion of consistency. Predictive objectives impose multiple forms of consistency across observations, which constrain the geometry of the representation space. Temporal consistency enforces continuity of events over time, contextual consistency stabilizes interpretations across varying inputs, and cross-modal consistency aligns representations derived from different modalities. Together, these constraints embed world knowledge as structural regularities rather than explicit transitions.

Large-scale self-supervised learning is critical for this process. Because training signals are derived directly from data, models can be exposed to vast and diverse corpora spanning text, images, video, and multimodal inputs. This diversity enables representations to internalize broad regularities of the world, often giving rise to emergent capabilities such as spatial reasoning, temporal inference, and commonsense understanding without explicit supervision \cite{Zong2025}.

Transformer architectures, introduced in \cite{Vaswani2017}, provide the computational foundation for this paradigm. Their attention-based structure enables modeling of long-range dependencies and supports flexible integration of multimodal inputs within a unified representation space. Combined with scalability in both data and model size, transformers have become the dominant architecture for implicit world models, including large language models, vision–language systems, and recent generative video models.

Taken together, predictive learning, consistency constraints, large-scale self-supervised training, and transformer architectures define the core learning paradigm of implicit world models. The following sections examine how these principles are realized in representative model classes.

\subsection{Language and Vision–Language Models as Implicit World Models}
Language and vision–language models (LLMs and VLMs) provide canonical examples of implicit world models. Trained on large-scale text and multimodal data with predictive objectives, they learn representations that encode regularities of the world without relying on explicit dynamics modeling. This demonstrates that substantial world knowledge can emerge in representations through predictive learning alone.

\textbf{Implicit world knowledge in VLMs:} Modern language models are trained using next-token prediction over large text corpora, where descriptions of events and interactions implicitly reflect how the world operates. Through this process, models learn to associate linguistic patterns with underlying structure, enabling them to generate coherent continuations that respect causal order, temporal progression, and contextual constraints. This capability is further enhanced in vision–language models, where representations are grounded in both linguistic and visual inputs, allowing more direct encoding of spatial structure and physical interactions.

A defining factor in this process is scale. Modern models are trained on massive datasets spanning hundreds of billions to trillions of tokens, often combining text, images, and video collected from diverse environments \cite{Zhao2023}. Such large-scale exposure enables representations to internalize broad regularities of the world, giving rise to emergent capabilities such as commonsense reasoning, spatial inference, and temporal understanding. In this sense, world knowledge in implicit models is not explicitly programmed but arises from the combination of predictive learning and large-scale data. This effect is further amplified by the scale of modern transformer models, which often contain hundreds of billions of parameters, providing sufficient capacity to internalize complex world structure.

\textbf{Inference and reasoning in VLMs:} The presence of world knowledge in VLMs becomes most evident at inference time. As illustrated in Figure 10, these models can infer plausible future events from partial observations in a manner analogous to human reasoning. Rather than simulating future rollouts through an explicit dynamics model, they perform inference over structured representations, generating candidate interpretations and selecting those that are consistent with learned regularities.

\begin{figure}[h]
  \centering
  \includegraphics[width=0.8\linewidth]{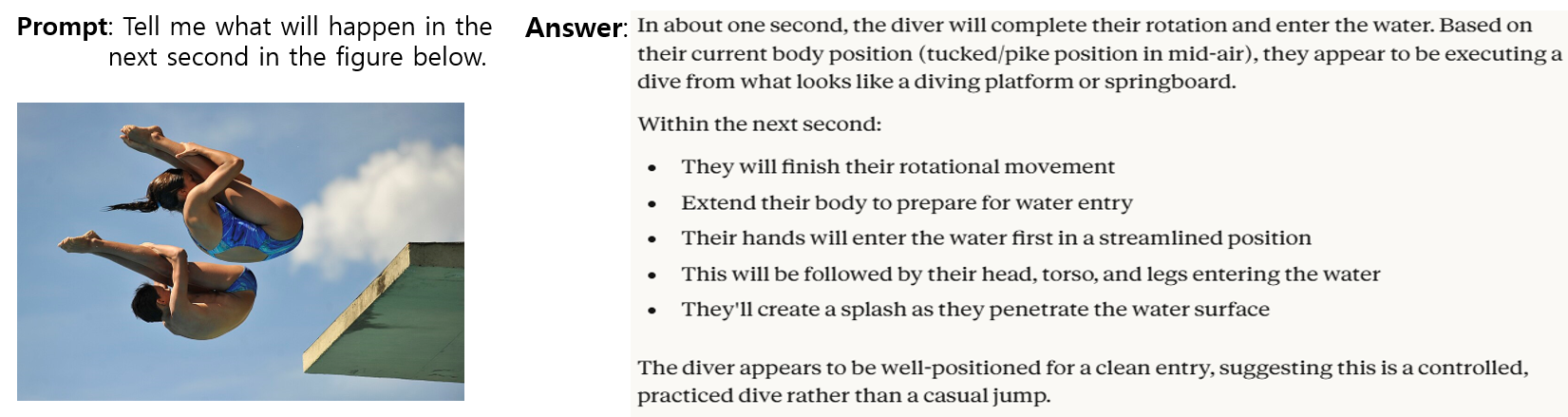}
  \caption{Example of human-like world knowledge and temporal reasoning exhibited by a VLM from a partial observation.}
  \Description{VLM}
\end{figure}

This representation-based inference manifests in several characteristic behaviors. Zero-shot generalization shows that learned representations contain reusable structure that can be applied to previously unseen tasks. Chain-of-thought prompting reveals how reasoning can emerge as a sequence of intermediate steps, in which interpretations are progressively refined to maintain consistency with world knowledge. In-context learning further demonstrates that models can rapidly adapt to new tasks by conditioning on a small number of examples, effectively forming local hypotheses within the representation space without updating parameters \cite{Wei2022, Bubeck2023}.

From the perspective of Figure 9(b), these behaviors can be interpreted as instances of the internal interaction loop: reasoning proceeds through successive refinement of representations rather than through rollout-based simulation. In this sense, inference in implicit world models corresponds to constraint-driven interpretation within a structured representation space.

\textbf{From representational knowledge to actionable systems:} While VLMs capture rich world knowledge in their representations, they are not explicitly grounded in action and reward during training. Unlike explicit world models, whose learning is centered around action–outcome relationships, implicit world models acquire knowledge through predictive learning without an explicit notion of action in the objective. Bridging this gap to enable actionable behavior therefore presents an additional challenge.

In practice, this gap is addressed by coupling pretrained representations with task-specific adaptation and interaction. Fine-tuning on downstream data introduces an external interaction loop, as illustrated in Figure 9(b), through which candidate actions proposed by representation-based inference are evaluated and refined using feedback from the environment. This process preserves the internal inference mechanisms of implicit world models while aligning them with the constraints of the target task.

Recent work explores this direction by integrating vision–language representations with perception and control modules, enabling action selection guided by inference over learned representations. We revisit this emerging paradigm in detail in Section V, where we examine how implicit world models can be extended to support physical interaction and control.

\subsection{Generative Video World Models: Genie}
Generative video models extend implicit world modeling to settings in which future observations can be synthesized directly from learned representations. Rather than relying on explicit dynamics models, these approaches learn to generate coherent future scenes by capturing regularities of the physical world from large-scale video data. A representative example is Genie \cite{Bruce2024}, which demonstrates how controllable, world-like behavior can emerge purely from observational learning without access to action labels or environment simulators.

\textbf{Video as a source of world knowledge:} Video data provides a rich and scalable source of supervision for learning world structure. Unlike static images or text, video naturally encodes temporal continuity, where object motion, physical interactions, and causal relationships unfold over time. As illustrated by the diver example in Figure 10, predicting the continuation of a video requires maintaining consistent body dynamics and interaction with the surrounding water. This consistency reflects underlying physical regularities—such as gravity and acceleration—that must be implicitly respected for the predicted motion to remain plausible.

Predictive objectives that require forecasting future frames therefore encourage models to internalize regularities such as persistence, motion consistency, and interaction constraints. At scale, learning from large and diverse video corpora enables these regularities to be embedded in latent representations, allowing models to generate coherent future observations without relying on explicit state-transition equations.

\textbf{Inference: autoregressive generation with latent control:} We begin with inference, as it most directly illustrates the generative behavior of the model. Figure 11 illustrates the inference-time operation of Genie. Given an initial observation \(x_1\) as a prompt, the encoder \(E\) maps it to a latent representation \(z_1\). A predictive model \(F\) then generates subsequent latent states autoregressively. At each step, the model predicts the next latent state \(z_{t+1}=F(z_t,a_t)\), where \(a_t\) denotes a latent control (action) variable that does not correspond to a physical action, but represents an abstract factor influencing the evolution of the scene. The decoder \(D\) maps these latent states back to image frames, yielding a sequence of plausible future observations.

\begin{figure}[h]
  \centering
  \includegraphics[width=0.8\linewidth]{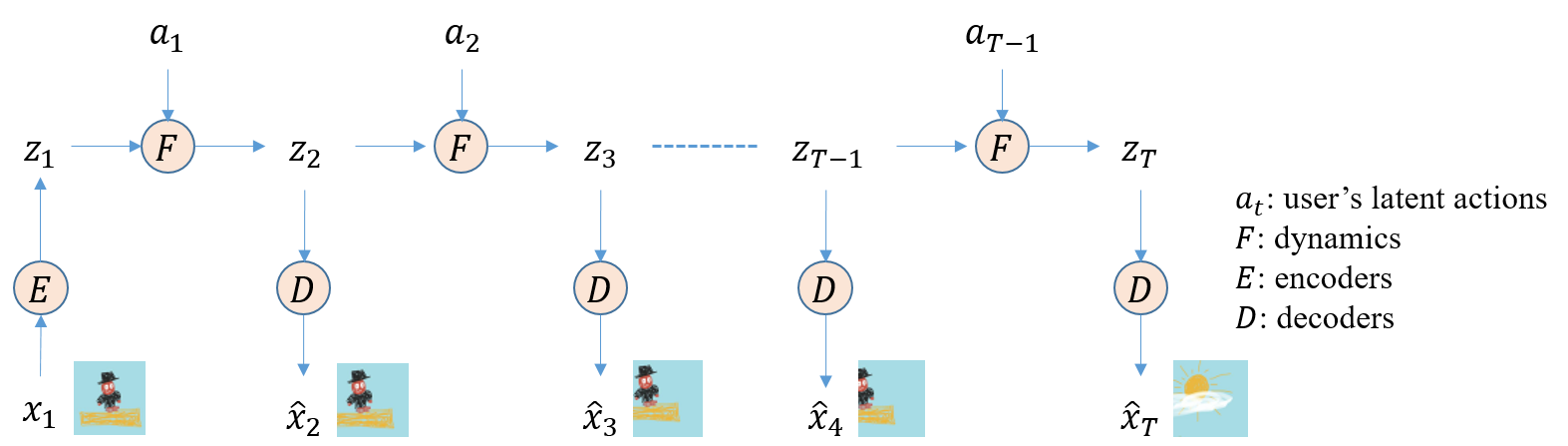}
  \caption{Inference-time autoregressive generation with latent control in Genie. Starting from an encoded frame, the latent dynamics model \(F\) recursively predicts future latent states conditioned on latent actions, while the decoder \(D\) reconstructs video frames autoregressively.}
  \Description{Genie-1}
\end{figure}

Although this procedure resembles the rollout mechanism used in explicit world models, its interpretation is fundamentally different. The predictive function \(F\) does not represent a physical dynamics model derived from an environment. Instead, it defines a generative transition in latent representation space, producing samples from a learned distribution of possible futures.

Latent control variables modulate the generation process, enabling multiple plausible rollouts to emerge from the same initial observation. While not directly interpretable as physical actions, their effects are consistent, allowing interactive exploration of the learned world dynamics.

\textbf{Learning: stage-wise training from video:} Figure 12 summarizes the training procedure used to learn Genie from raw video observations. The model is trained in a stage-wise manner that first establishes a stable perceptual representation and then learns controllable latent dynamics.

\begin{figure}[h]
  \centering
  \includegraphics[width=0.7\linewidth]{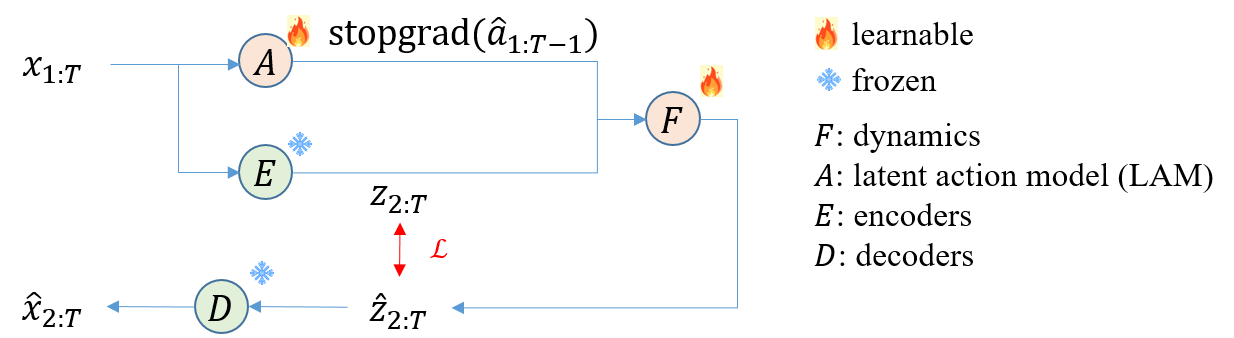}
  \caption{Stage-wise training procedure of Genie from video data. The encoder \(E\) and decoder \(D\) are first pretrained for video reconstruction, after which the latent action model \(A\) and latent dynamics model \(F\) are trained to predict future latent states autoregressively in latent space.}
  \Description{learning Genie-1}
\end{figure}

In the first stage, a video tokenizer is trained using a VQ-VAE architecture. The encoder \(E\) maps video frames \(x_t\) into discrete latent tokens \(z_t\), and the decoder \(D\) reconstructs frames from these tokens. This stage provides a compact and stable representation of visual observations.

In the second stage, the encoder \(E\) and decoder \(D\) are frozen, and the model learns latent control variables \(a_t\) that explain temporal changes in the video sequence. These variables are inferred directly from data without external supervision and represent controllable factors of variation in the observed dynamics. At the same time, the predictive model \(F\) is trained to model the latent transition \(z_{t+1}=F(z_t,a_t)\). Training proceeds by minimizing prediction error between predicted latent representation \(\hat{z}_{2:T}\) and observed latent representations \(z_{2:T}\). Because the learning objective enforces consistency with real video sequences, the model gradually captures temporal regularities such as motion patterns, interaction constraints, and object persistence.

Through this process, Genie learns representations that support both prediction and controllable generation. World knowledge is thus embedded in the interaction between latent states \(z_t\), latent controls \(a_t\), and the predictive model \(F\), rather than in explicit state-transition equations or action-labeled supervision.

\subsection{World Models without Generation: Joint-Embedding Predictive Architectures}
The generative models discussed in the previous section demonstrate that explicit simulation is not required for learning world structure. Joint-embedding predictive architectures (JEPA) take this idea one step further by removing generation altogether \cite{LeCun2022}. Instead of synthesizing observations, JEPA learns world structure purely through predictive alignment in representation space. In this framework, world modeling is reframed from generating future observations to determining which representations are compatible with the current context. World knowledge is thus encoded not as explicit dynamics or generative processes, but as constraints within the geometry of the representation space.

\textbf{Principles:} JEPA learn world structure by modeling compatibility relationships between representations derived from different views of the same underlying data. As illustrated in Figure 13(a), two inputs \(x\) and \(y\), sampled from a common scene or event, are encoded into representations \(s_x=E_x(x)\) and \(s_y=E_y(y)\). A predictor network estimates a target representation as \(\hat{s}_y=P(s_y)\), and learning proceeds by minimizing an energy or distance function between \(\hat{s}_y\) and \(s_y\).

\begin{figure}[h]
  \centering
  \includegraphics[width=0.8\linewidth]{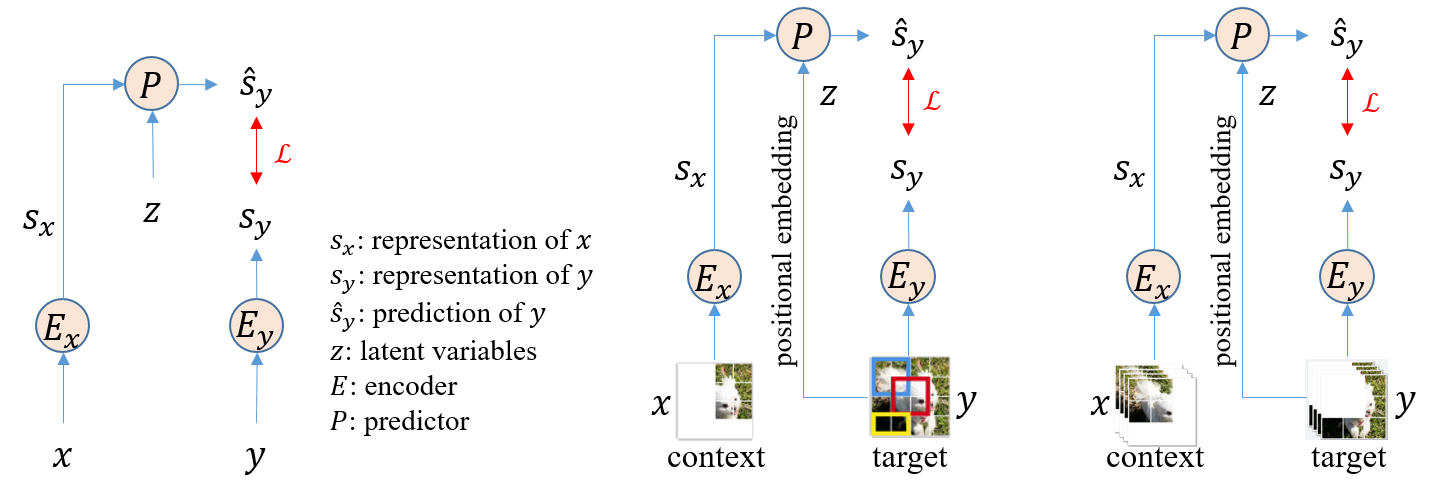}
  \caption{Joint-embedding predictive architectures (JEPA). (a) Generic JEPA learns compatibility in representation space via energy-based prediction. (b) I-JEPA performs spatial prediction in images without reconstruction. (c) V-JEPA performs temporal prediction in video, capturing future-compatible representations as an implicit world model.}
  \Description{various JEPA}
\end{figure}

At its core, JEPA implements a self-supervised pattern-completion objective in representation space. Given partial information, the model predicts a representation that is consistent with the underlying data rather than reconstructing observations. Because both \(x\) and \(y\) originate from the same scene or event, their representations share spatial, temporal, or semantic structure, and the learning objective encourages \(\hat{s}_y\) to be compatible with \(s_y\).

In its more general formulation, JEPA introduces a latent variable \(z\), yielding \(\hat{s}_y=P(s_y,z)\), to account for multiple compatible targets. This latent variable captures unobserved factors—such as viewpoint, temporal evolution, or object motion—that give rise to multiple plausible representations, thereby representing uncertainty in the underlying data. By varying \(z\), the model can represent a set of compatible futures without explicitly generating observations.

Learning is driven by an energy function defined in representation space, where compatible context–target pairs are assigned low energy and incompatible pairs higher energy. In this way, the geometry of the learned representation space encodes world knowledge as compatibility constraints. World modeling is thus reframed from predicting what the world will look like to determining which representations are consistent with the current context. This stands in contrast to reconstruction-based approaches such as masked autoencoders \cite{He2022}, which aim to recover missing observations, whereas JEPA focuses on learning structural compatibility rather than sensory detail.

\textbf{I-JEPA: learning spatial structure from images \cite{Assran2023}:} Figure 13(b) illustrates the application of the JEPA framework to images. Given an input image, a subset of regions is selected as context, while disjoint regions are treated as targets. The context regions are encoded into a representation \(s_x\), and a predictor network estimates representations of the target regions.

The prediction is performed entirely in representation space: the model learns to align predicted target representations with those obtained from the corresponding image regions, without reconstructing pixels. This formulation implements spatial pattern completion, where the model infers which representations are compatible with the observed context.

To specify where predictions should be made, the predictor is provided with location information indicating the positions of the target regions. In practice, this information is often implemented using positional embeddings, as illustrated in Figure 13(b), but its role is purely to indicate target locations rather than to encode semantic content. Location information specifies where predictions are made, whereas the latent variable \(z\) in the general JEPA formulation captures uncertainty about which compatible representation should be selected. 

Through this process, I-JEPA learns structural properties of images—such as object layout, part–whole relationships, and contextual dependencies—by modeling compatibility across spatial regions. Empirical results show that the learned representations capture meaningful spatial structure and support downstream tasks such as classification and segmentation, indicating that I-JEPA encodes world knowledge beyond low-level appearance.

\textbf{V-JEPA: learning temporal structure from video \cite{Bardes2024}:} Figure 13(c) extends the JEPA framework to video, where the goal is to model temporal structure by predicting representations of future or unobserved segments from partial observations. Given a video sequence, a subset of spatio-temporal regions is treated as context, while the remaining regions serve as targets. The context is encoded into a representation \(s_x\), and a predictor estimates representations of the target segments.

As in I-JEPA, prediction is performed entirely in representation space without reconstructing video frames. The model learns to align predicted representations with those of the target segments, implementing temporal pattern completion. This enables the model to capture which future representations are compatible with the observed past, rather than generating explicit pixel-level predictions.

To indicate where predictions should be made, the predictor is provided with location information specifying the positions of the target regions in space and time. In practice, this may be implemented using positional encodings, but its role is solely to index target locations. As in the general JEPA formulation, a latent variable \(z\) can be introduced to account for multiple compatible futures, capturing uncertainty in temporal evolution.

Through this formulation, V-JEPA learns temporal regularities such as object persistence, motion continuity, and event progression. Empirical results show that the learned representations support downstream tasks, including action recognition and video understanding, indicating that V-JEPA captures meaningful temporal world knowledge without relying on explicit dynamics modeling or generative video synthesis.

\textbf{Implications and limitations of non-generative world models:} The JEPA framework positions world modeling as learning compatibility relationships in representation space, rather than generating observations or simulating rollouts. This perspective complements existing approaches: explicit world models enable planning through simulation, generative models explore plausible futures through sampling, and JEPA constrains these futures by capturing structural consistency.

However, this design also introduces limitations. Because JEPA does not perform action-conditioned prediction or explicit rollout, it cannot directly evaluate the consequences of candidate actions or support planning through internal simulation. As a result, it is less suited for tasks that require precise, action-dependent forecasting. In practice, extending JEPA to such settings requires integrating additional components that connect representation learning with action and control.

This representation-centric perspective also underlies recent extensions to embodied systems, such as V-JEPA 2 for robotics and AD-L-JEPA for autonomous driving, which will be discussed in the next section.

\section{Physical AI}
Chapters III and IV examined how agents learn world models, either explicitly or implicitly. Physical AI extends this perspective to embodied systems operating in the real world, where deployment conditions differ fundamentally from simulation. The physical environment introduces uncertainty, friction, latency, hardware degradation, wind disturbances, surface variations, and partial observability. This mismatch—commonly referred to as the \textit{sim-to-real gap}—implies that accurate prediction alone is insufficient; agents must anticipate how their actions unfold under imperfect and evolving conditions, where errors are costly and data collection is limited.

\subsection{From World Modeling to Physical Intervention}
Recent research in physical AI is shifting from achieving peak performance on narrowly defined tasks toward developing systems capable of generalizing across diverse tasks and environments\cite{Fung2025}. This transition toward task generality exposes the limitations of purely reactive approaches, which rely on direct perception–action mappings and struggle to adapt beyond their training conditions. As a result, internal world models are becoming increasingly important, providing a structured mechanism for anticipating action outcomes, reusing knowledge across tasks, and supporting long-horizon reasoning. In this section, we formalize this transition through the Mode-1/Mode-2 framework and examine its implications for physical AI.

\textbf{Motivations and issues:} Most existing studies of world models have been conducted in virtual environments, such as games or simulated robotics, where interaction is safe, data is abundant, and dynamics are well controlled. However, if world models are to serve as a foundation for general intelligence, they must extend beyond virtual settings and operate in the physical world.

Physical AI systems operating in real-world environments face several fundamental challenges. First, the sim-to-real gap—arising from distribution shift, partial observability, and environmental uncertainty—limits the transfer of models trained in simulation. Second, physical interaction is inherently expensive, requiring high data efficiency and effective reuse of experience. Third, real-world deployment imposes strict constraints on real-time operation and safety. Finally, agents must reason about how their actions influence future states under imperfect and evolving conditions, requiring tight coupling between perception, action, and prediction.

In this setting, two distinct paradigms of decision making emerge (Figure 14), formalized by the Mode-1/Mode-2 distinction for the perception–action loop \cite{LeCun2022, Kahneman2011}. Reactive physical AI (Mode-1) directly maps perception to action through short-horizon feedback loops, corresponding to association and intervention in the reasoning hierarchy (Figure 2). In contrast, world model–based physical AI (Mode-2) introduces an internal world model that supports planning and reasoning by enabling agents to evaluate action-conditioned future rollouts before execution, as illustrated in Figure 14. This shift reflects a transition toward imagination, where decisions are guided not only by immediate sensory input but also by internally simulated outcomes. While reactive systems are effective for narrowly defined tasks requiring speed and reliability, advancing toward general intelligence requires the capabilities of Mode-2.

\begin{figure}[h]
  \centering
  \includegraphics[width=0.8\linewidth]{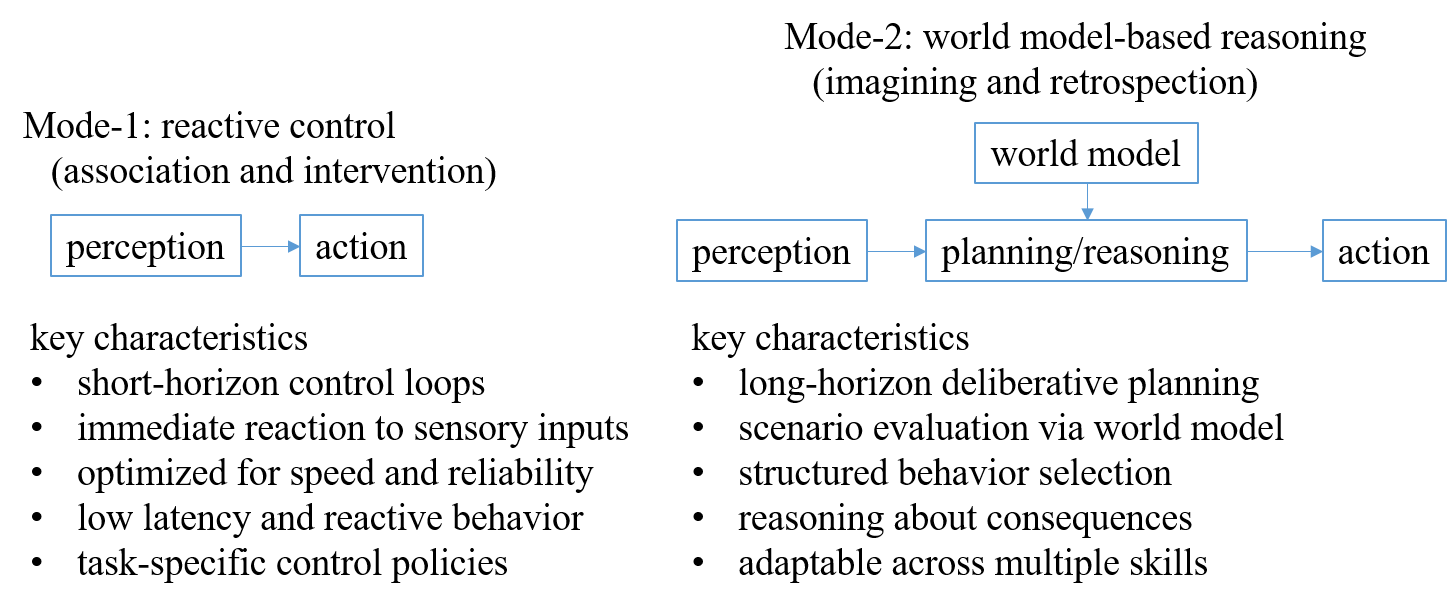}
  \caption{Reactive (Mode-1) and world model-based (Mode-2) physical AI.}
  \Description{Mode-1 vs. Mode-2}
\end{figure}

\textbf{Reactive physical AI (Mode 1):} Reactive physical AI refers to a class of systems in which actions are selected directly from current observations through a policy that maps state \(s_t\) to action \(a_t\). This mapping may be engineered, as in classical control systems, or learned through data-driven approaches such as RL. In either case, behavior is determined primarily by the present state, without explicitly evaluating alternative future outcomes or maintaining an internal model of environment dynamics, making it inherently suited for task-specific applications.

A defining characteristic of reactive physical AI is its reliance on short-horizon perception–action coupling, as illustrated in Figure 14. Decisions are made based on immediate sensory input, enabling fast and stable responses that are well suited for time-critical tasks. These properties make reactive approaches particularly effective in real-world deployment, where low latency, robustness, and reliability are essential.

This paradigm underlies a wide range of real-world systems, from industrial automation and household robotics to emerging humanoid platforms. Recent systems demonstrate increasingly sophisticated sensorimotor capabilities in challenging physical settings, including agile humanoid whole-body control \cite{He2025}, high-speed interactive tasks such as humanoid table tennis \cite{Su2025}, and highly dynamic flight in insect-scale robots \cite{Hsiao2025}.

However, because reactive systems do not explicitly reason about future trajectories, they are inherently limited in supporting long-horizon planning, compositional behavior, and adaptation to novel or changing environments. These limitations motivate the transition toward world model–based approaches.

\begin{figure}[h]
  \centering
  \includegraphics[width=0.8\linewidth]{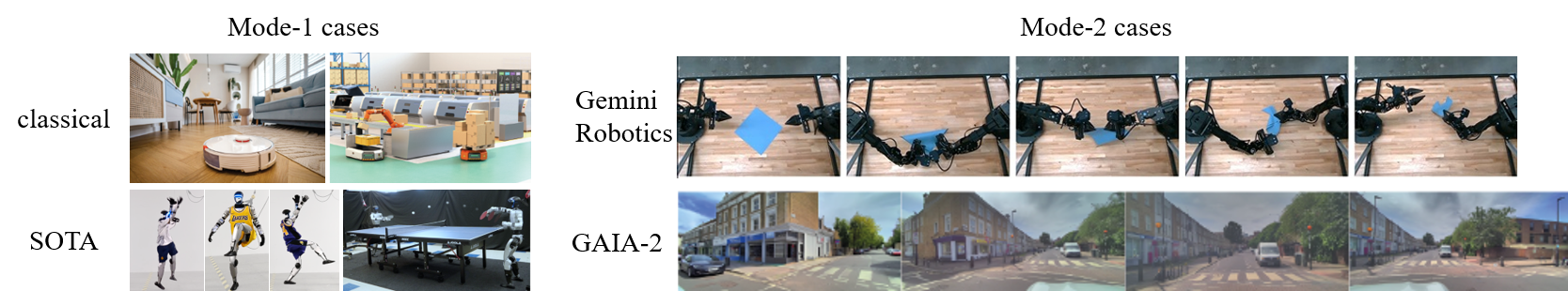}
  \caption{Examples of Mode-1 and Mode-2 physical AI systems. Mode-1 (reactive) systems include commercially deployed and task-specialized platforms such as household robots and industrial automation, as well as emerging humanoid systems relying on fast perception–action loops. Mode-2 (world model–based) systems, including robotics and autonomous driving, evaluate action-conditioned futures through internal models to support planning and long-horizon decision making.}
  \Description{Mode-1 and Mode-2 examples}
\end{figure}

\textbf{World model-based physical AI (Mode-2):} World model–based physical AI refers to systems that select actions by reasoning over predicted future outcomes, rather than reacting solely to current observations. In contrast to reactive approaches, Mode-2 maintains an internal latent state that summarizes past observations and actions, and uses a learned world model to anticipate how the environment evolves under candidate interventions. This enables evaluation of action-conditioned future rollouts before execution, supporting planning and long-horizon decision making, as illustrated in Figure 14.

A defining characteristic of Mode-2 is the tight coupling between perception, prediction, and action. Given a latent state, the agent generates future rollouts using the world model and selects actions based on their predicted consequences. This may be realized through explicit rollouts or predictive representations that implicitly encode future outcomes, shifting decision making from reactive response to anticipatory control.

Formally, let \(s_t\) denote the latent state at time \(t\), which summarizes past observations and actions and can be interpreted as a belief state under partial observability. A world model \(f\) predicts future states conditioned on actions by \(s_{t+1}=f(s_t,a_t)\), and can be recursively applied to evaluate candidate action sequences. A policy \(\pi\) selects actions based on these predictions with respect to a task-dependent objective. This separation is essential: the world model captures environment dynamics independently of specific tasks, while the policy determines behavior according to task objectives.

This structure enables long-horizon planning, compositional behavior, and adaptation, thereby providing a key mechanism for achieving task generalization across diverse environments. Because the world model encodes general dynamics rather than task-specific mappings, it can be reused across multiple tasks, improving data efficiency and facilitating transfer.

Recent advances build on this principle by leveraging large-scale datasets and sequence modeling architectures, particularly transformers, to learn generalizable predictive structures. In this setting, world models are often implemented as sequence models that unify perception, prediction, and action within a common representation space. The following sections examine how these ideas are realized in real-world systems, including robotics and autonomous driving.

\subsection{World Model-Based Physical AI in Robotics}
We now examine how world model–based physical AI is realized in robotics systems. In particular, we focus on two representative paradigms—explicit latent world models and implicit predictive models—and analyze how they support generalizable behavior in real-world settings.

\subsubsection{Trends and Issues}
Recent advances in robotics have been accompanied by a rapid expansion in the diversity of robot platforms, operational environments, and task objectives. Physical robots operate with heterogeneous sensory modalities, including proprioception, force feedback, and high-dimensional visual inputs such as RGB and depth images. This multimodal and temporally structured input, together with the variability of real-world environments, favors sequence modeling architectures such as transformers, which can capture long-range dependencies and cross-modal interactions. As a result, modern robotic systems are increasingly deployed in unstructured and dynamic settings, performing tasks that span manipulation, navigation, and whole-body control.

This growing diversity has made task generalization a central challenge. Systems optimized for narrowly defined tasks often fail to transfer across new objects, environments, or task specifications, revealing the limitations of task-specific policies and reactive control. Enabling robots to generalize across tasks and conditions has therefore become a primary objective in physical AI.

Recent progress toward this goal has been driven by large-scale robotic datasets and world model–based approaches. Datasets such as DROID \cite{Khazatsky2024} and Open X-Embodiment \cite{Padalkar2023} provide diverse, cross-task and cross-platform data, supporting learning beyond narrowly defined settings. Building on this data, a range of approaches has emerged, including explicit latent world models such as DayDreamer \cite{Wu2022}, implicit predictive models such as V-JEPA 2 \cite{Assran2025}, and foundation model–based systems such as COSMOS \cite{Nvidia2025} and Octo \cite{Ghosh2024}. In parallel, vision–language–action (VLA) systems, including Gemini Robotics \cite{GeminiRobotics2025}, extend this paradigm by incorporating language-conditioned action, enabling more flexible and goal-directed behavior \cite{Firoozi2023}.

These developments point toward a unifying perspective in which large-scale data and world models enable generalizable robotic intelligence. In the following sections, we examine representative approaches along the axis of explicit and implicit world models.

\subsubsection{Explicit Latent World Models: DayDreamer}
DayDreamer provides an early real-world realization of world model–based physical AI, extending the Dreamer architecture from simulated environments to robotic systems \cite{Wu2022}. As illustrated in Figure 16, DayDreamer preserves the same latent world model structure as Dreamer, differing primarily in that data is collected from real-world interaction rather than simulation. This demonstrates that explicit world model learning, based on latent dynamics and imagination-based rollouts, can be directly extended from simulation to physical systems.

\begin{figure}[h]
  \centering
  \includegraphics[width=0.8\linewidth]{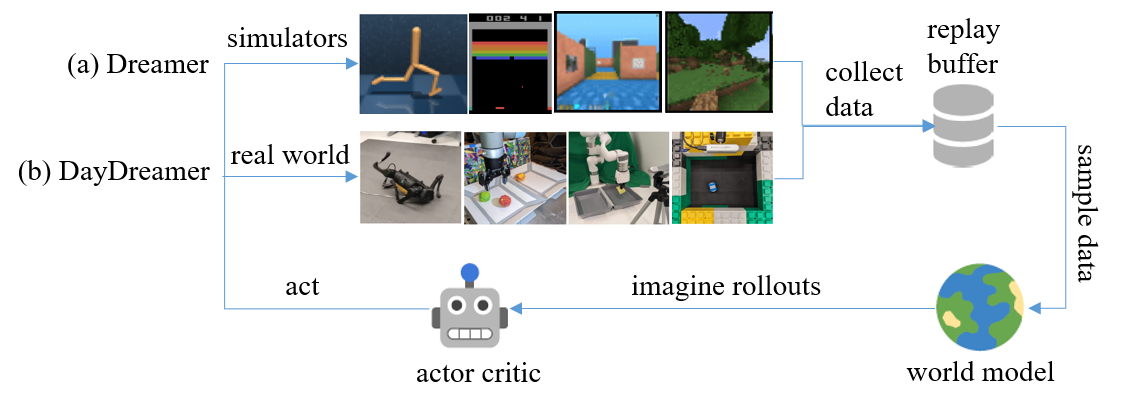}
  \caption{Dreamer vs. DayDreamer: Extension of explicit latent world model learning from simulation in (a) to real-world interaction in (b) with identical model structure.}
  \Description{DayDreamer}
\end{figure}

In DayDreamer, multi-modal sensory inputs are encoded into a stochastic latent state representation following the recurrent state-space model (RSSM) formulation. The world model predicts action-conditioned latent transitions over a planning horizon, and imagined rollouts are generated through latent states. These imagined rollouts are used to optimize an actor–critic policy with respect to a reward-driven objective. This explicit formulation enables direct evaluation of action-conditioned future outcomes through simulation in latent space.

Crucially, while the world model captures task-agnostic embodied dynamics—such as joint motion, contact interactions, and object behavior—the policy becomes task-specific by optimizing a particular reward function. This separation allows a shared world model to support multiple tasks, while policies specialize to task objectives.

Unlike its simulation-based predecessor, DayDreamer operates under strict real-world constraints. Data collection is slow and costly, exploration must remain safe to avoid hardware damage, and control must be stable under real-time execution. As a result, training must be carefully designed to ensure sample efficiency, robustness, and safety.

DayDreamer is evaluated across multiple robotic platforms, including manipulation arms, quadruped locomotion systems, and mobile robots, and across diverse tasks such as manipulation, navigation, and whole-body control. Despite this diversity, the system employs a shared neural architecture and largely consistent hyperparameters, demonstrating that a single latent world model can support multiple task-specific behaviors. These results provide empirical evidence that reusable physical knowledge, combined with reward-driven policy optimization, offers a viable foundation for scalable and generalizable embodied intelligence.

This explicit, rollout-based approach contrasts with implicit predictive models, introduced in the following section, which capture future structure in representation space without performing explicit simulation.

\subsubsection{Implicit Predictive World Models: V-JEPA 2}
While DayDreamer realizes model-based control through explicit latent rollouts, V-JEPA 2 represents an alternative paradigm that learns predictive structure directly in latent space through large-scale representation learning, without performing explicit rollouts \cite{Assran2025}. As illustrated in Figure 17, V-JEPA 2 is first pretrained on large-scale internet video data using a joint-embedding predictive objective. This stage learns representations that capture general visual and temporal structure, enabling the model to encode predictive relationships without reconstructing raw observations. The pretrained model is then adapted to downstream robotic tasks using a relatively small amount of robot interaction data.

\begin{figure}[h]
  \centering
  \includegraphics[width=0.8\linewidth]{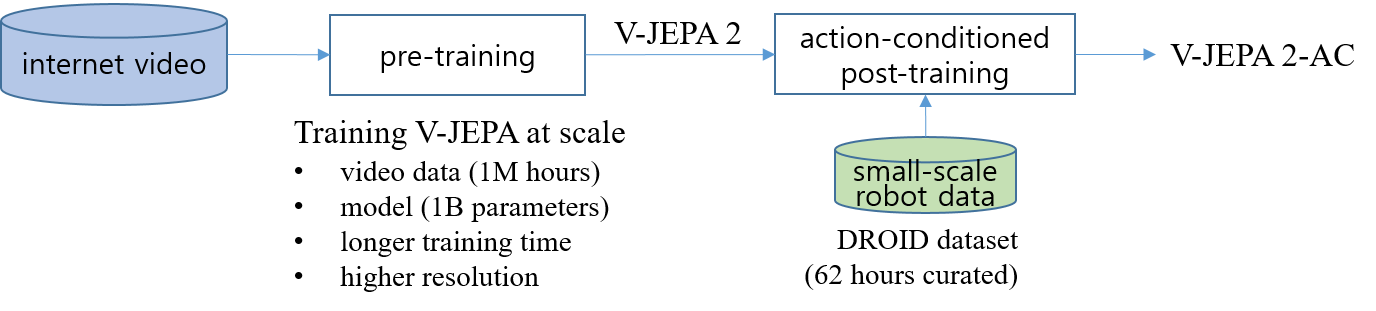}
  \caption{Pretraining and adaptation pipeline of V-JEPA 2. Representations are learned from large-scale internet video data and adapted to robotics using small-scale robot interaction data, illustrating the data-efficient nature of implicit predictive world models. (adapted from \cite{Assran2025}).}
  \Description{V-JEPA 2}
\end{figure}

For robotics, this adaptation is realized in V-JEPA 2-AC (Action-Conditioned), as illustrated in Figure 18, where a pretrained visual encoder is combined with action-conditioned prediction. Given a sequence of observations \(x_{\leq k}\), the model is trained to predict the representation of a future frame \(y=x_{k+1}\) conditioned on both past observations and actions. Specifically, the model encodes the past observations \(x_{\leq k}\) into a latent representation \(s_x\) using a pretrained encoder \(E_x\), and a predictor \(P\) estimates a future representation \(\hat{s}_y\) conditioned on both the latent state and action-related inputs. The target frame \(y\) is encoded into \(s_y\) by a second encoder \(E_y\), and training minimizes a discrepancy between \(\hat{s}_y\) and \(s_y\). This formulation follows the JEPA principle in Figure 13, where prediction is performed in latent space without reconstructing raw observations. Here, \(z\) denotes action-related signals, such as robot actions and poses, which can be derived from differences in end-effector states between adjacent frames.

\begin{figure}[h]
  \centering
  \includegraphics[width=0.8\linewidth]{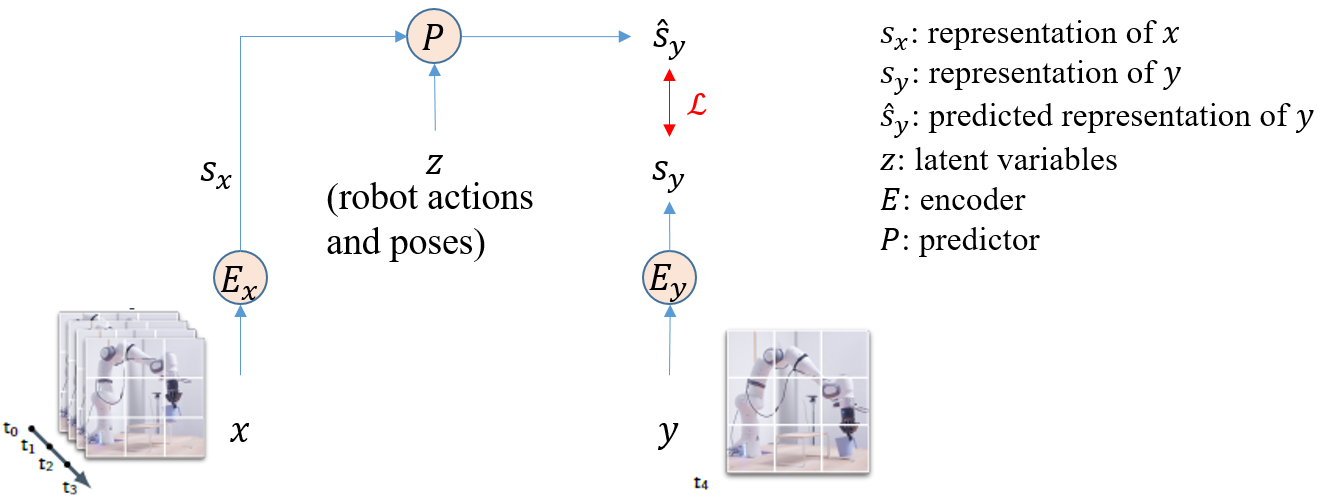}
  \caption{Fine-tuning of V-JEPA 2-AC (adapted from \cite{Assran2025}). Frozen pretrained encoders \(E_x\) and \(E_y\)
generate visual representations, while the predictor \(P\) is trained to predict future representations conditioned on robot actions and poses.}
  \Description{V-JEPA 2}
\end{figure}

At inference time, control is achieved through goal-conditioned planning in representation space, which can be interpreted as a form of model predictive control (MPC). Given a current observation and a goal image, candidate action sequences are evaluated by predicting their resulting representations and minimizing their discrepancy to the goal representation. Planning is performed in a receding-horizon manner, executing only the first action before re-optimizing at the next step.

Despite its scalability and data efficiency, this approach has several limitations. Planning is constrained by finite horizons and iterative optimization, and the reliance on goal images—rather than language or abstract task specifications—limits flexibility. These limitations highlight both the promise and current boundaries of implicit predictive world models in robotics.

\subsubsection{Discussion: Explicit vs Implicit in Robotics}
The two approaches discussed above—explicit latent world models and implicit predictive models—represent fundamentally different design paradigms for world model–based physical AI in robotics. Explicit models, exemplified by DayDreamer, learn action-conditioned dynamics and evaluate candidate behaviors through latent rollouts, enabling planning through internal simulation of candidate actions. In contrast, implicit models such as V-JEPA 2 learn predictive structure in representation space without constructing explicit rollouts, relying instead on representation-level prediction for planning.

This distinction leads to complementary strengths and limitations. Explicit world models offer greater controllability and interpretability, as imagined rollouts can be directly inspected and optimized, but they typically require substantial task-specific interaction data and careful system design to ensure stability and efficiency. Implicit models, by contrast, scale more naturally with large datasets, leveraging internet-scale pretraining and requiring only limited robot interaction for adaptation, as illustrated in Figure 17. However, their reliance on representation-level prediction constrains planning flexibility, particularly in long-horizon or abstract task settings.

In practice, these paradigms are not mutually exclusive but reflect different trade-offs between structure and scalability. Explicit models are well suited for settings where precise control and task-specific optimization are critical, while implicit models offer a promising path toward generalizable robotic intelligence through large-scale learning. Understanding and combining these complementary properties remains an important direction for future research in physical AI.

\subsection{World Model-Based Physical AI in Autonomous Driving}
Autonomous driving is one of the most demanding real-world applications of physical AI. While commercial robotaxi services already operate in selected urban regions, most systems remain at Level 4 autonomy, limited to geofenced areas supported by high-definition (HD) maps. Achieving robust performance under extreme weather, rare traffic configurations, and unforeseen events remains an open challenge \cite{Zhao2025, Kusano2025}.

The difficulty stems from the open-world nature of driving, characterized by multi-agent interactions, long temporal horizons, and significant uncertainty. Rare long-tail events—though infrequent—are often safety-critical. Addressing these challenges requires moving beyond reactive control toward systems capable of anticipating the consequences of actions before execution through internal predictive modeling.

\subsubsection{Trends and Issues}
Figure 19 illustrates the architectural evolution of autonomous driving systems. Early approaches relied on modular pipelines that decomposed driving into perception, prediction, planning, and control. Environmental knowledge—such as HD maps, lane graphs, and kinematic constraints—was externalized as hand-engineered world models, enabling structured reasoning but limiting generalization beyond predefined domains. End-to-end methods sought to reduce engineering complexity by directly mapping sensor inputs to control commands using large-scale imitation learning, sometimes augmented with RL \cite{Zhao2025}. However, these approaches remain fundamentally reactive, as decisions are derived from instantaneous observations without explicit evaluation of future scenarios.

\begin{figure}[h]
  \centering
  \includegraphics[width=0.8\linewidth]{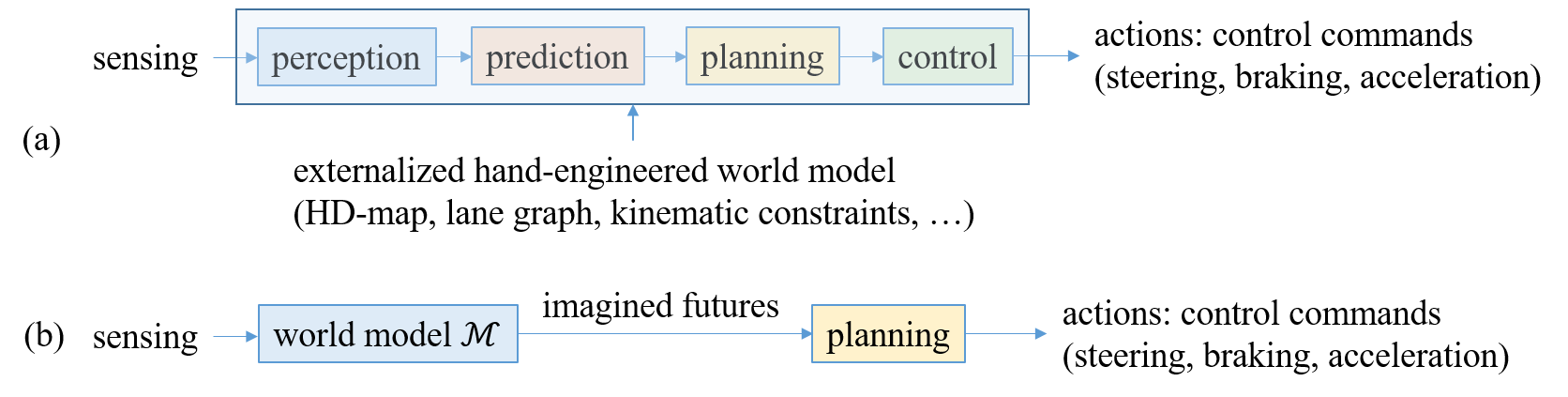}
  \caption{Architectural evolution in autonomous driving. (a) Traditional modular pipelines based on hand-engineered world models. (b) World model-based driving with internal predictive modeling and planning.}
  \Description{Autonomous driving paradigms}
\end{figure}

Unlike robotics, there is no canonical explicit latent world model that cleanly supports action-conditioned rollouts. The complexity of driving environments—including multi-agent interactions, partial observability, and high-dimensional sensory inputs—makes such formulations difficult to realize in practice. As a result, recent approaches tend to lie along a spectrum between explicit generative world models and implicit predictive models, rather than adhering to a single unified formulation.

The lower part of Figure 19 reflects a shift toward internal predictive world modeling. Instead of relying on externalized structure or purely reactive policies, the system learns an internal model that generates imagined futures under candidate actions and evaluates their consequences through planning. Action selection is thus guided by internally predicted rollouts rather than immediate observations.

This transition—from externalized representations and reactive control to internal predictive modeling—enables counterfactual reasoning in open-world environments. In autonomous driving, where safety depends on anticipating rare but critical events, such capability is essential, particularly for addressing long-tail scenarios that are difficult to capture through limited real-world experience \cite{Zhang2026}. The growing body of work on world models for driving \cite{Guan2024, Gao2025} further reflects this emerging paradigm.

\subsubsection{Generative World Models: GAIA}
GAIA-1 exemplifies an explicit generative instantiation of internal world modeling for autonomous driving \cite{Hu2023}. As illustrated in Figure 20, the system models scene dynamics by predicting future latent tokens in an autoregressive manner, enabling the generation of future rollouts in latent space under multimodal conditioning.

\begin{figure}[h]
  \centering
  \includegraphics[width=0.6\linewidth]{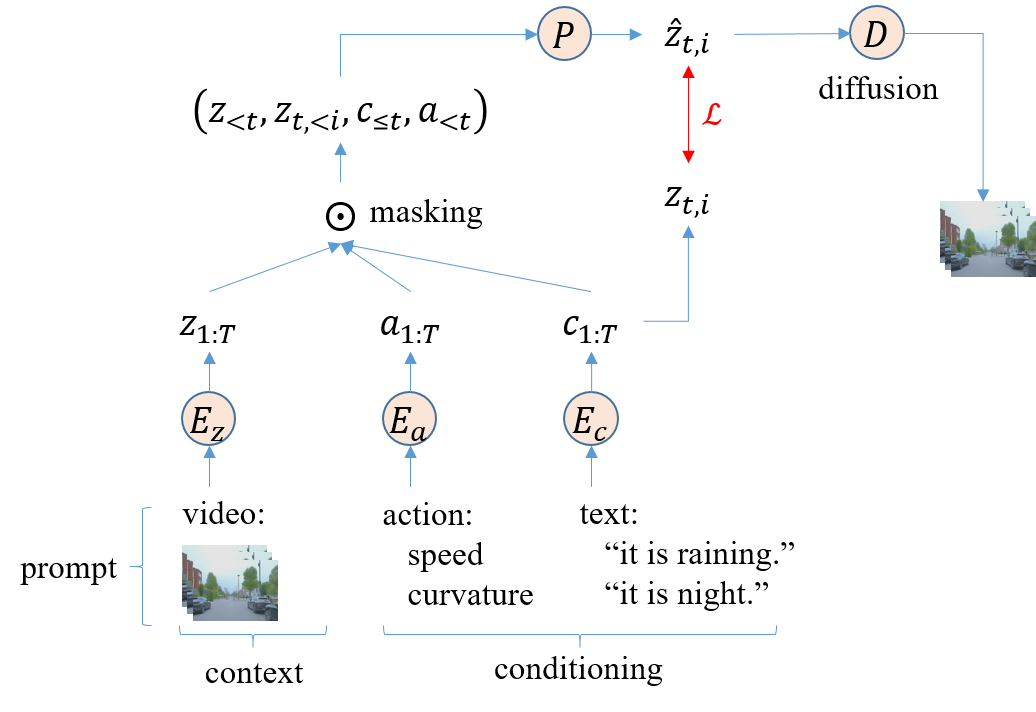}
  \caption{Simplified architecture of GAIA-1. The world model is realized as an autoregressive predictor in latent token space, while a diffusion-based decoder reconstructs video frames for visualization (adapted from \cite{Hu2023}).}
  \Description{GAIA}
\end{figure}

Given a context video together with auxiliary inputs such as actions (e.g., speed and curvature) and textual descriptions (e.g., weather or time of day), the inputs are encoded into a sequence of latent tokens representing the spatiotemporal structure of the scene. These tokens serve as the state for the world model. The model then predicts each latent token \(\hat{z}_{t,i}\) conditioned on past tokens \(z_{<t}\), previously generated tokens within the same frame \(z_{t,<i}\), and conditioning inputs \(c_{\leq t}\) and \(a_{<t}\), as shown in Figure 20.

The core world model is realized as a transformer-based autoregressive predictor \(P\), which generates future latent tokens sequentially. Training minimizes a negative log-likelihood loss over the predicted tokens, corresponding to maximizing the likelihood of the observed latent sequence. This formulation enables the model to learn action-conditioned dynamics, multi-agent interactions, and scene evolution directly in latent space.

Crucially, this architecture constitutes an explicit world model: future scene evolution is generated through autoregressive rollout of latent tokens. Unlike implicit predictive models that estimate future representations without simulation, GAIA explicitly constructs future rollouts, enabling direct evaluation of candidate scenarios under different conditions.

In the full system, a diffusion-based decoder \(D\) can optionally reconstruct video frames from predicted latent tokens; however, this component serves primarily as a rendering layer for visualization or supervision and is not part of the core world model. At inference time, the autoregressively generated latent tokens \(\hat{z}_{t,i}\) can be optionally decoded into video frames using this decoder for visualization or downstream processing.

Subsequent developments, including GAIA-2 \cite{Russell2025} and GAIA-3 \cite{Wayve2025}, further explore controllability, multi-view consistency, and large-scale deployment for safety evaluation. These efforts indicate that generative world modeling is evolving into a broader paradigm for scalable and robust autonomous driving systems.

\subsubsection{Implicit Predictive World Models: AD-L-JEPA}
AD-L-JEPA extends the JEPA to autonomous driving by operating on structured spatial representations derived from BEV (Bird’s Eye View) point clouds \cite{Zhu2025}. Instead of explicit reconstruction or rollout, it performs prediction directly in latent space, learning representations that capture scene structure.

As shown in Figure 21, a BEV point cloud is partitioned into context and target via structured masking. The context retains visible regions, while the target contains masked regions to be inferred. These inputs are encoded by \(E_x\) and \(E_y\), producing latent representations \(s_x\) and \(s_y\). The predictor \(P\) maps the context representation to \(\hat{s}_y\), which is compared with \(s_y\) using a loss computed only over masked regions.

\begin{figure}[h]
  \centering
  \includegraphics[width=0.6\linewidth]{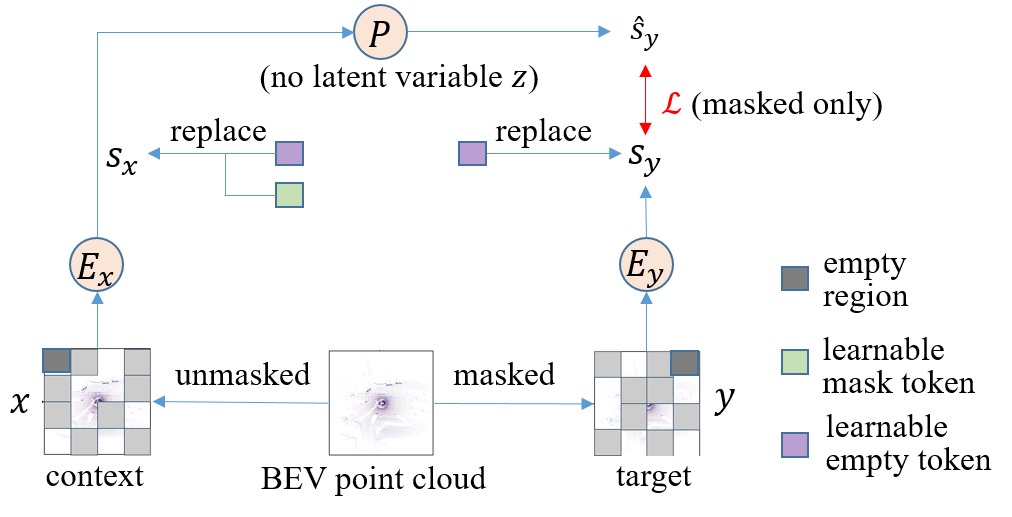}
  \caption{Overall architecture of AD-L-JEPA, where masked BEV regions are predicted in latent space using learnable token replacement.}
  \Description{AD-L-JEPA}
\end{figure}

A central component is the token replacement mechanism applied after encoding. In both \(s_x\) and \(s_y\), empty regions (cells without LiDAR points) are replaced with a learnable empty token, and masked regions with a learnable mask token. The modified representation \(s_x\) is then fed into \(P\), ensuring that the input explicitly encodes which regions are observed, absent, or to be inferred. These learnable tokens are essential for stable learning. The empty token provides a consistent representation for regions without observations, preventing noisy or trivial outputs over large empty areas. The mask token explicitly indicates prediction targets, guiding the model to infer missing structure. Without these tokens, the model cannot reliably distinguish between observed, empty, and masked regions, leading to ill-posed learning and degraded performance.

Unlike the original JEPA formulation, AD-L-JEPA does not introduce an explicit latent variable z; prediction is deterministic, and compatibility between \(\hat{s}_y\) and \(s_y\) is enforced via a regression loss in latent space. By computing loss only on masked regions, the model focuses on inferring unseen structure rather than reproducing observed inputs.

AD-L-JEPA can be viewed as a spatial extension of I-JEPA to structured 3D environments. While I-JEPA predicts masked image regions and V-JEPA predicts future frames, AD-L-JEPA applies the same principle to BEV grids, capturing spatial relationships relevant to autonomous driving. This provides a unified view of implicit world models as predictive representation learners across modalities.
Finally, AD-L-JEPA is primarily a representation learning framework rather than a full decision-making system. Its effectiveness is demonstrated in downstream tasks such as 3D object detection, where learned representations improve performance over training from scratch. In contrast to generative models such as GAIA, which simulate future observations, AD-L-JEPA operates entirely in latent space without decoding or rollout, representing a distinct paradigm of implicit world modeling.

\subsubsection{Discussion: Explicit vs Implicit in Autonomous Driving}
The two approaches presented in this section—generative world models such as GAIA and predictive implicit models such as AD-L-JEPA—highlight a fundamental distinction in how world knowledge is represented and utilized in autonomous driving. Explicit models learn environment dynamics that can be directly simulated, enabling rollout-based reasoning and planning. In contrast, implicit models encode world structure within latent representations and perform prediction without constructing explicit future trajectories.

This distinction leads to complementary strengths and limitations. Generative models provide a powerful framework for long-horizon reasoning and counterfactual analysis, but incur substantial computational cost and may introduce compounding errors through rollout. Implicit models, by operating directly in latent space, offer greater efficiency and scalability, and can leverage large-scale data for representation learning. However, they lack explicit mechanisms for multi-step planning and structured exploration of future scenarios.

These observations suggest that explicit and implicit world models should not be viewed as competing paradigms, but as complementary approaches. In autonomous driving, generative models can support planning and simulation, while implicit models provide robust and scalable representations for perception and downstream tasks. A key direction for future research lies in integrating these paradigms, combining the efficiency of implicit representations with the reasoning capabilities of explicit world models.

\subsection{Foundation Models for Physical AI: Toward Generalist Embodied Intelligence}
Foundation models are large-scale models trained on diverse data to learn general-purpose representations that can be adapted to a wide range of downstream tasks \cite{Bommasani2021}. In the context of physical AI, they aim to capture broad world knowledge that extends beyond task-specific world models.

A representative example is Cosmos, which is trained on large-scale video data spanning diverse domains \cite{Nvidia2025}. By learning from extensive visual experience, Cosmos captures general patterns of object dynamics, interactions, and environmental structure, providing a shared prior that can be transferred across tasks such as robotics and autonomous driving. In this sense, it can be viewed as a world foundation model that generalizes the role of world models beyond individual tasks.

Another important direction is exemplified by Gemini Robotics, which builds upon large-scale vision–language models and extends them to embodied settings \cite{GeminiRobotics2025}. Rather than learning world knowledge from scratch, such systems leverage pre-trained foundation models and adapt them to perception and action through additional multimodal and interaction data. This enables robots to perform complex tasks by combining general knowledge with task-specific adaptation.

These developments suggest a natural progression from world models to foundation models. While traditional world models are often trained for specific environments or tasks, foundation models aim to capture general-purpose world knowledge that can be reused across domains. From this perspective, explicit and implicit world models can be seen as complementary building blocks: explicit models provide structured reasoning through simulation, while implicit models offer scalable representation learning from large data.

Looking forward, a key challenge is to integrate these capabilities within unified systems for physical AI. Such systems would combine the ability to learn general world knowledge at scale with the capacity for reasoning, prediction, and action in real-world environments, moving toward more general and adaptive forms of intelligence.

\section{Pathways and Challenges toward AGI}
The developments discussed throughout this tutorial suggest that world models provide a promising foundation for advancing toward artificial general intelligence (AGI) \cite{Hendrycks2025}. By enabling agents to represent, predict, and reason about the structure of the world, both explicit and implicit world models support more generalizable and data-efficient intelligence than purely reactive approaches. Recent progress in large-scale pretraining, predictive representation learning, and physical AI further indicates that world modeling is evolving from task-specific prediction toward broader forms of world understanding. However, significant challenges remain before such systems can achieve human-level intelligence and autonomy. The key pathways and challenges toward AGI can be summarized as follows:

\begin{itemize}
    \item \textbf{Unified world modeling.} A central direction toward AGI is the integration of explicit and implicit world modeling. Explicit models provide structured dynamics and support rollout-based reasoning, planning, and counterfactual evaluation. Implicit models, in contrast, encode rich world knowledge within scalable representations learned from large-scale data. Future systems are likely to combine these complementary strengths, integrating structured prediction with representation learning to achieve both reasoning capability and scalability. Such systems must also remain grounded in physical interaction and real-world feedback, enabling robust adaptation across diverse environments.
    \item \textbf{Hierarchical prediction and planning.} Human intelligence operates across multiple levels of abstraction and time scales, ranging from low-level motor control to long-term reasoning and planning. Current world models demonstrate promising capabilities in short-horizon prediction and planning, but remain limited in hierarchical reasoning across extended temporal horizons. While short-term prediction can rely on low-level sensory representations, long-horizon reasoning requires increasingly abstract representations that suppress irrelevant details while preserving high-level structure. Achieving AGI may therefore require hierarchical world models capable of multi-scale prediction, task decomposition, and abstract planning across different levels of abstraction and temporal scales \cite{LeCun2022}.
    \item \textbf{Intention and goal formation.} Finally, current models lack an explicit notion of intention and goal formation, which are central to human intelligence. Existing systems can optimize predefined objectives or follow external instructions, but they generally do not possess intrinsic mechanisms for generating, selecting, and adapting goals based on context. Human intelligence is characterized not only by the ability to predict and act, but also by the capacity to autonomously form intentions, prioritize objectives, and flexibly adjust behavior in response to changing environments. Addressing this limitation requires moving beyond reactive or externally guided systems toward agents capable of internally constructing and reasoning about goals, representing a critical step toward AGI.
\end{itemize}

\section{Conclusions}    
This tutorial has presented world models as a unifying framework for intelligent systems, distinguishing between explicit models that support rollout-based reasoning through structured dynamics and implicit models that encode predictive structure within scalable learned representations. Together, these complementary paradigms provide a foundation for physical AI, enabling prediction-driven intelligence beyond reactive control in domains such as robotics and autonomous driving, while recent foundation models suggest a pathway toward unified systems integrating perception, prediction, and action.

Despite rapid progress in world modeling and foundation models, major gaps remain between current systems and human intelligence. Existing models still struggle to construct abstract predictive representations across multiple temporal scales, perform robust long-horizon reasoning, and autonomously generate and adapt goals in open-ended environments. Addressing these limitations will require advances beyond scaling alone, including hierarchical world modeling, embodied interaction, and architectures capable of integrating prediction, planning, and autonomous objective formation.

In summary, world modeling is emerging not merely as a component of intelligent systems, but as a central principle for understanding and constructing adaptive intelligence. Although current systems remain far from achieving AGI, the convergence of world modeling, foundation models, and embodied learning suggests a plausible pathway toward more general, scalable, and autonomous intelligence. More broadly, these developments increasingly resonate with biologically inspired views of intelligence in which hierarchical prediction, world-centered representation, and embodied interaction constitute fundamental organizing principles of cognition.
    
\begin{acks}
\end{acks}

\bibliographystyle{ACM-Reference-Format}
\bibliography{sample-base}
\end{document}